\documentclass[11pt]{article}

\PassOptionsToPackage{table}{xcolor}
\usepackage[final]{acl}

\usepackage{times}
\usepackage{latexsym}
\usepackage[T1]{fontenc}
\usepackage[utf8]{inputenc}
\usepackage{placeins}
\usepackage{microtype}

\usepackage{inconsolata}

\usepackage{graphicx}
\usepackage{amsmath}
\usepackage{cleveref}
\usepackage{booktabs}
\usepackage{multirow}
\usepackage{amsfonts}
\usepackage{array}
\usepackage{float}
\usepackage{subcaption}
\title{EarlySelectKV: Anticipatory KV Selection via Cross-Layer Query Proxies}

\author{
  \textbf{Yuhao Liu\textsuperscript{1,2}},
  \textbf{Yuehao Li\textsuperscript{1,2}},
  \textbf{Linghe Zhang\textsuperscript{1,2}},
  \textbf{Gaobin Huang\textsuperscript{1,2}} \\
  \textbf{Junran Li\textsuperscript{1,2}},
  \textbf{Hongchang Wang\textsuperscript{1}},
  \textbf{Haifang Jian\textsuperscript{1}}\thanks{Corresponding author:
  \href{mailto:jhf@semi.ac.cn}{\texttt{jhf@semi.ac.cn}}.} \\[2pt]
  \normalfont\normalsize
  \textsuperscript{1}\enspace Laboratory of Solid State Optoelectronics Information Technology, \\
  Institute of Semiconductors, Chinese Academy of Sciences, Beijing, China \\
  \textsuperscript{2}\enspace University of Chinese Academy of Sciences, Beijing, China \\[2pt]
  \texttt{\{liuyuhao,liyh,zhanglinghe23,huanggaobin23\}@semi.ac.cn} \\
  \texttt{\{lijunran23,hcwang,jhf\}@semi.ac.cn}
}

\begin{document}
\maketitle
\begin{abstract}
Query-conditioned sparse KV selection reduces long-context inference cost by attending to a subset of historical KV entries at each decoding step. Existing selectors must wait for the target-layer query before scoring candidates and generating indices, placing routing on the immediate pre-attention path. We introduce \textsc{EarlySelectKV}, a training-free transformation that uses proxy queries from earlier hidden states to precompute selected KV indices while retaining the real target-layer query for final attention. Tests with RocketKV, Quest, and Loki show no consistent pattern of quality degradation across the evaluated settings. In the primary RocketKV evaluation, ES-RocketKV differs by \(+0.018\) points on LongBench, approximately \(-0.01\) points on RULER, and \(+0.25\) percentage points on representative NIAH/Passkey settings. In a TensorRT-LLM deployment with CPU-offloaded routing, ES-RocketKV improves end-to-end output throughput by up to 12.4\% and reduces GPU-resident memory by 11.1\% of the combined KV/K-summary pool. This makes routing timing an explicit optimization target alongside KV subset selection in sparse long-context decoding.\footnote{Code is available at \url{https://github.com/sea-with-sakura/EarlySelectKV}.}
\end{abstract}

\section{Introduction}

Long-context inference in large language models requires substantial memory capacity and bandwidth to store and access historical key-value (KV) states \citep{PagedAttention,sparQ}. The KV cache grows linearly with context length, batch size, and model depth, making these costs especially relevant to long-input workloads such as long-document understanding and repository-level code completion \citep{LongBench,RepoBench}.

\begin{figure}[t]
    \centering
    \includegraphics[width=\linewidth]{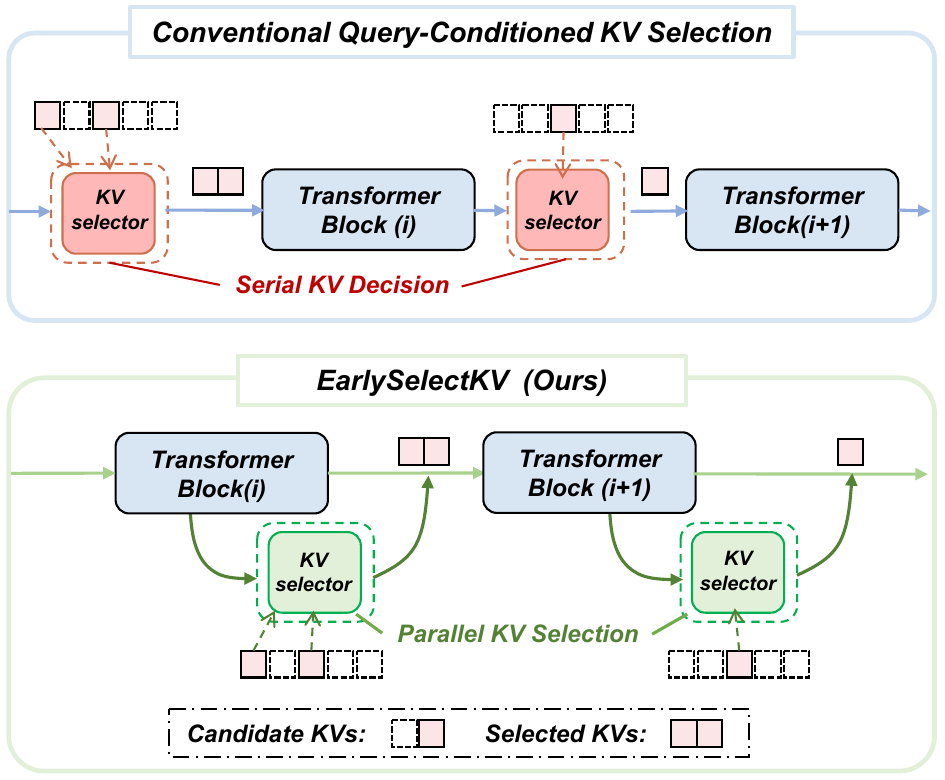}
    \caption{
    Conventional query-conditioned KV selection and \textsc{EarlySelectKV}.
    Conventional selectors wait for the target-layer query, so index generation remains on the immediate pre-attention path. EarlySelectKV derives a proxy query from preceding hidden states and precomputes KV indices in an earlier computation window.
    }
    \label{fig:intro_compare}
\end{figure}

A common strategy reduces the number of historical KV entries read by attention. Eviction and compression methods maintain a compact cache by retaining recent tokens, attention sinks, or tokens estimated to be important from prompt-time observations \citep{streamLLM,H2O,SnapKV}. Dynamic sparse methods instead keep a candidate cache and select a small KV subset at each decoding step according to the current query. This query-conditioned interface adapts to changing evidence needs during generation, but the selected-token count alone does not determine its cost \citep{QUEST,sparQ,loki}.

As shown in Figure~\ref{fig:intro_compare}, conventional query-conditioned selectors must obtain the target-layer query before scoring candidate pages or tokens, producing selected indices, and preparing the corresponding KV entries. The resulting serial routing dependency precedes sparse attention. Reducing the sparse budget lowers attention computation, but routing computation and metadata access remain exposed on the same critical path. Once attention is sparsified, deciding which KV entries to access can become a visible source of latency.

The optimization target is therefore not only which KV entries to keep, but also when their indices can be determined. Routing and value aggregation require different levels of query fidelity. Attention over the selected KV states needs the exact target-layer query, whereas routing only needs to preserve the relative preference among candidates.

We propose EarlySelectKV, a training-free transformation for query-conditioned sparse KV selectors. It advances index generation without changing sparse attention: the proxy query is used only for routing, while the real query still performs attention over the original selected KV states. We integrate EarlySelectKV with representative query-conditioned sparse KV selectors, including RocketKV, Quest, and Loki, demonstrating its plug-and-play applicability across different routing designs. Rather than redesigning the selector, EarlySelectKV advances the availability of selected KV indices, making query-conditioned sparsity more amenable to system-level scheduling.

As a representative case, ES-RocketKV matches RocketKV on LongBench (+0.018 points), Needle-in-a-Haystack (+0.25 percentage points), and RULER (-0.01 points), indicating negligible accuracy degradation. This parity makes the selected indices a reliable scheduling signal. In a deployment with CPU-offloaded routing, the early availability of selected indices allows routing and metadata preparation to be staged before the target attention step, reducing exposed stalls on the GPU attention path and lowering GPU-resident auxiliary state.

Our contributions are summarized as follows:

\begin{itemize}
    \item We propose EarlySelectKV, a training-free framework for anticipatory KV-cache selection. It uses query anchors from preceding hidden states to reduce routing overhead on the critical path of dynamic sparse decoding.

    \item EarlySelectKV provides a selector-preserving interface for query-conditioned routing. It retains the original candidate construction and scoring rules while supporting anchor choices with different lookahead windows and query-projection approximations.

    \item Experiments on LongBench, Needle-in-a-Haystack, and RULER show that EarlySelectKV maintains task performance and selection fidelity while reducing the visible routing overhead of exact-query routing.
\end{itemize}

\begin{figure}[t]
    \centering
    \includegraphics[width=\linewidth]{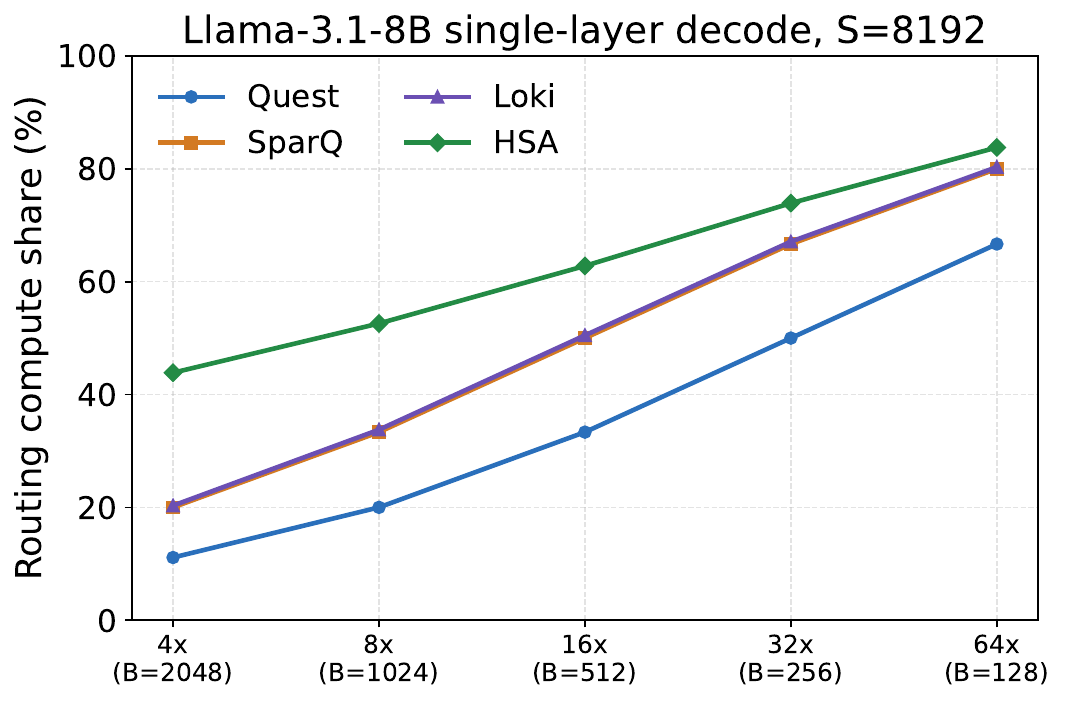}
    \caption{
    Theoretical routing overhead under sparse attention budgets.
    We analyze a single Llama-3.1-8B layer with an 8K KV cache and vary the selected-token budget.
    As the budget decreases, sparse attention becomes cheaper, and routing accounts for a larger fraction of the remaining attention-side computation.
    }
    \label{fig:routing_overhead}
\end{figure}

\begin{figure*}[t]
    \centering
    \includegraphics[width=0.98\linewidth]{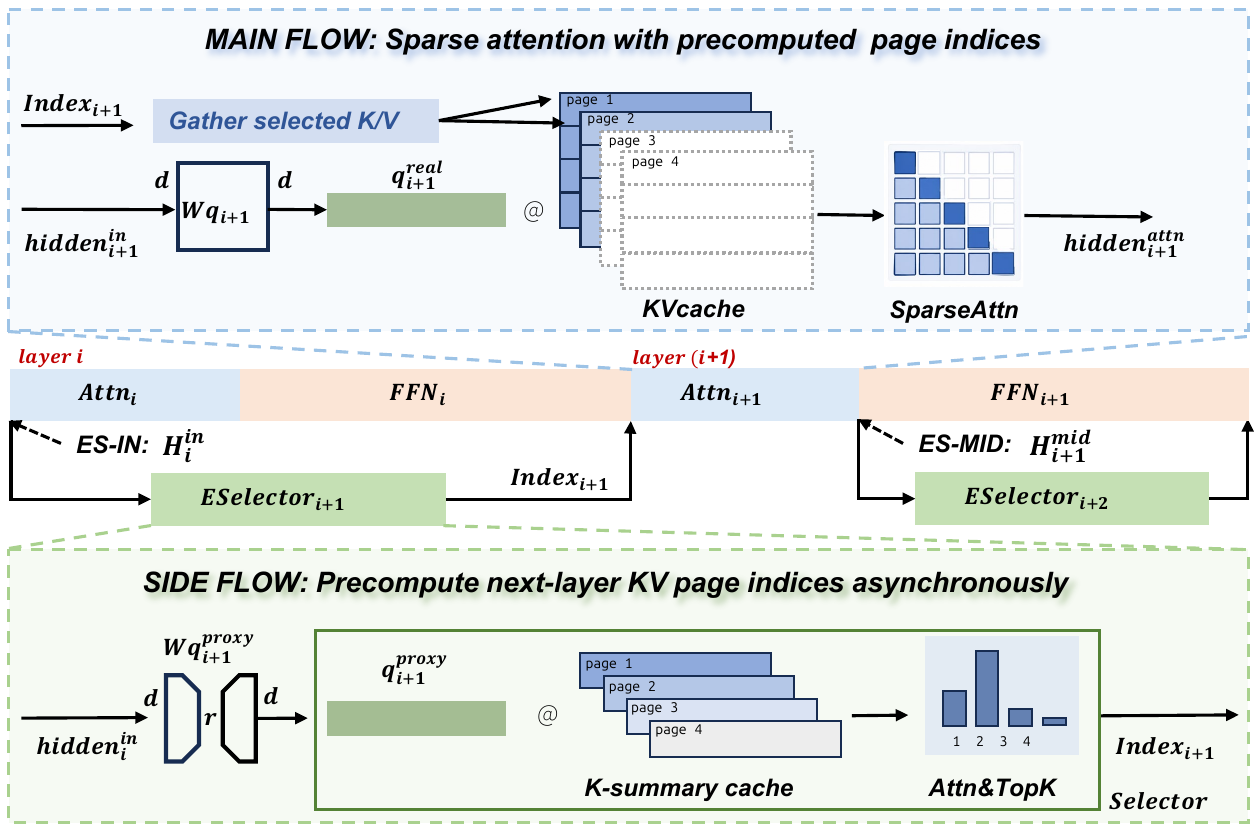}
    \caption{
    Overview of EarlySelectKV. The main flow gathers selected KV using precomputed page indices and performs sparse attention with the real query. The side flow constructs a proxy query from an earlier hidden state and precomputes next-layer page indices. Early and Mid taps represent two timing--fidelity trade-offs: the former offers a longer preparation window, while the latter provides higher selection fidelity.
    }
    \label{fig:earlyselectkv_overview}
\end{figure*}

\section{Related Work}

\paragraph{Architecture-level KV-Cache Efficiency.}
Architectural methods reduce KV-cache cost by changing how KV states are represented, shared, or reused.
MQA~\citep{MQA} and GQA~\citep{GQA} share KV states across query heads, whereas MLA~\citep{MLA} compresses them into latent representations.
Other designs reuse KV states across layers~\citep{brandon2024kvsharing} or combine local and global attention patterns~\citep{YOCO}.
Trainable sparse-attention designs such as InfLLM-V2~\citep{InfLLMV2} switch between dense and block-sparse attention for long-context adaptation.
Their savings apply to every decoding step and can benefit dense as well as sparse inference, since the reduction is built into how attention stores or reuses historical states.
However, their applicability is coupled to the model architecture and often requires training-time integration or specialized deployment support.

\paragraph{KV-Cache Eviction and Offloading.}
Without modifying the model architecture, inference-time methods reduce memory pressure through KV eviction or CPU--GPU placement.
H2O~\citep{H2O} retains recent tokens and heavy hitters, while Scissorhands~\citep{Scissorhands} exploits persistent token importance; StreamingLLM~\citep{streamLLM} preserves attention sinks and a recent window; and SnapKV~\citep{SnapKV} compresses prompt KV using an observation window.
Lookahead Q-Cache~\citep{LAQ} incorporates anticipated query information into eviction through pseudo lookahead queries constructed before generation.
These methods reduce memory footprint, but their retained history is largely fixed after pruning, offering limited per-step adaptivity.
For CPU-offloaded inference, InfiniGen~\citep{InfiniGen} uses cross-layer speculation to prefetch important KV entries, whereas ScoutAttention~\citep{ScoutAttention} predicts next-layer queries to perform CPU-side sparse attention one layer ahead.

\paragraph{Query-Conditioned Sparse KV Access.}
Query-conditioned sparse access retains candidate KV states and selects or samples an active subset separately for each decoding query.
Quest~\citep{QUEST} uses page-level query bounds, SparQ~\citep{sparQ} exploits query-channel sparsity, Loki~\citep{loki} uses low-rank scoring, and MagicPIG~\citep{MagicPIG} applies query-dependent LSH sampling.
TokenSelect~\citep{TokenSelect} selects tokens per head; RetrievalAttention~\citep{RetrievalAttention} uses vector retrieval.
RocketKV~\citep{RocketKV} is a hybrid method that combines prompt-stage KV eviction with query-conditioned coarse-to-fine sparse attention.
Because the active subset can change at each step, these methods preserve decoding-step adaptivity and can maintain accuracy under small per-query KV-access budgets.
This flexibility nevertheless entails a timing dependency: most selectors cannot score candidates or materialize selected indices until the target-layer query becomes available, leaving both operations on the immediate pre-attention path.
Unlike IndexCache~\citep{IndexCache}, which reuses top-\(k\) indices from preceding layers, EarlySelectKV advances fresh layer-specific index generation using a proxy query while retaining the real target-layer query for final attention.

\section{Method}
\label{sec:method}

\subsection{Observation}
\label{sec:observation}

Long-context sparse decoding reduces inference cost by limiting the KV entries accessed by attention, but query-conditioned selectors add a routing step before sparse attention.
This step reads auxiliary routing states, such as page metadata or summary caches, and performs index-generation operations such as page scoring or top-\(k\) selection.
Although these operations do not participate in value aggregation, they gate sparse attention by producing the selected KV indices, placing routing-dependent work on the immediate pre-attention path.

Figure~\ref{fig:routing_overhead} illustrates this routing cost on the attention path.
As the selected-token budget decreases, sparse attention becomes cheaper, but routing need not shrink at the same rate.
Consequently, routing dominates an increasing fraction of the attention-path latency.
This suggests that, once attention is sparsified, deciding which KV entries to access can become a non-negligible latency component.

Existing work mainly addresses this issue by making routing cheaper or more accurate.
In contrast, we focus on when routing is performed.
Exact-query routing must wait until the target-layer query is produced, so the selected KV indices become available only on the immediate pre-attention path.
This motivates making the indices available earlier while preserving the behavior of the original selector.

\subsection{Overview}
\label{sec:method_overview}

EarlySelectKV is an anticipatory routing framework for query-conditioned sparse KV selectors.
As shown in Figure~\ref{fig:earlyselectkv_overview}, EarlySelectKV decomposes dynamic KV selection into a main flow and a side flow.
The main flow performs sparse attention at the target layer: it reads precomputed page indices, gathers the corresponding selected KV entries, and computes attention with the real query of the target layer.
The side flow runs before the main flow reaches the target attention layer. It constructs a proxy query from a preceding hidden state, performs lightweight selection over a K-summary cache or another routing view, and precomputes the page indices required by the next layer. This moves page-index generation off the immediate dependency path to target attention.

At decoding step \(t\), consider current layer \(\ell\) and the next layer \(\ell+1\). Given the real query \(q_{t,\mathrm{real}}^{\ell+1}\) of the next layer, a query-conditioned sparse selector can be abstracted as
\begin{equation}
    \mathcal{I}_t^{\ell+1}
    =
    \mathcal{R}_{\ell+1}
    \left(
        q_{t,\mathrm{real}}^{\ell+1},
        \mathcal{M}_t^{\ell+1};
        B
    \right),
    \label{eq:standard_selector}
\end{equation}
where \(\mathcal{R}_{\ell+1}\) denotes the underlying routing rule, \(\mathcal{M}_t^{\ell+1}\) is the routing view used by the selector, and \(B\) is the routing budget.
The routing view \(\mathcal{M}_t^{\ell+1}\) abstracts the selector's searchable state, ranging from the full KV cache to compact proxies such as candidate caches or summaries.
In standard execution, \(\mathcal{I}_t^{\ell+1}\) cannot be produced until \(q_{t,\mathrm{real}}^{\ell+1}\) becomes available; consequently, KV gathering and sparse attention are both blocked by this routing step.

EarlySelectKV replaces the real query in Equation~\ref{eq:standard_selector} with a proxy query constructed earlier:
\begin{equation}
    \widehat{\mathcal{I}}_t^{\ell+1}
    =
    \mathcal{R}_{\ell+1}
    \left(
        q_{t,\mathrm{proxy}}^{\ell+1},
        \mathcal{M}_t^{\ell+1};
        B
    \right).
    \label{eq:early_selector}
\end{equation}
This transformation changes only the query signal that drives routing, allowing the required page indices to be prepared before attention consumes the corresponding KV entries. The selector structure remains unchanged, and accuracy is maintained empirically in the evaluated settings.

\subsection{Proxy Query Settings}

For a compatible query-conditioned selector, EarlySelectKV wraps it into an anticipatory selection module, which we call ESelector. Given an anchor state \(\tilde{H}_{t,\ell}\) from the current layer, ESelector first constructs a proxy query for the next layer \(\ell+1\):
\begin{equation}
    q_{t,\mathrm{proxy}}^{\ell+1}
    =
    W_{q,\mathrm{proxy}}^{\ell+1}
    \operatorname{LN}_{\ell+1}(\tilde{H}_{t,\ell}),
    \label{eq:proxy_query}
\end{equation}
and then invokes the original selector to produce the precomputed indices.
Here, \(W_{q,\mathrm{proxy}}^{\ell+1}\) can be the native query projection \(W_q^{\ell+1}\) of the next layer or its low-rank approximation.
The native projection gives a closer match to the target-layer query space, whereas the low-rank variant lowers the cost of proxy-query construction.
Both choices are training-free.
Appendix~\ref{app:rank_tradeoff} further analyzes how the approximation rank trades proxy-query construction cost for task performance.

The choice of \(\tilde{H}_{t,\ell}\) determines ESelector's trade-off between preparation window and selection fidelity. We study two fixed anchors, ES-In and ES-Mid, and use offline layer-sensitivity profiling to derive a model-specific layer-wise assignment between them.

\paragraph{ES-In.}
ES-In feeds ESelector with an earlier block input.
At the current layer \(\ell\), ES-In uses \(H_{\ell}^{\mathrm{in}}\) to construct \(q_{t,\mathrm{proxy}}^{\ell+1}\) and precompute \(\widehat{\mathcal{I}}_t^{\ell+1}\).
Because this anchor is available before layer \(\ell\) is computed, ES-In provides a longer preparation window for routing, metadata lookup, and KV preparation.
It is suitable for layers that tolerate earlier proxy queries, especially when latency hiding is the main goal.

\paragraph{ES-Mid.}
ES-Mid uses a later intermediate hidden state as the ESelector input.
For a pre-LN Transformer block, the post-attention state is
\begin{equation}
    H_{\ell}^{\mathrm{mid}}
    =
    H_{\ell}^{\mathrm{in}}
    +
    \operatorname{Attn}_{\ell}
    \big(
        \operatorname{LN}_{\ell}(H_{\ell}^{\mathrm{in}}),
        K^{\ell},
        V^{\ell}
    \big),
\end{equation}
which already incorporates attention-level contextual information through the residual connection.
Specifically, ES-Mid uses \(H_{\ell}^{\mathrm{mid}}\) to construct \(q_{t,\mathrm{proxy}}^{\ell+1}\) and precompute \(\widehat{\mathcal{I}}_t^{\ell+1}\).
ES-Mid therefore provides a stronger proxy-query representation than ES-In, but leaves a shorter preparation window.
It is preferable for layers that are more sensitive to proxy-query approximation.
We therefore use ES-Mid as the conservative default.

\paragraph{ES-Mix.}
ES-Mix uses offline calibration to choose a fixed anchor for each layer.
Insensitive layers use ES-In to maximize the preparation window, whereas sensitive layers use ES-Mid to preserve selector fidelity.

Let \(M_{\ell+1}(Q_a)\) denote the real-query attention mass at layer \(\ell+1\) retained by the indices selected using proxy query \(Q_a\), averaged over calibration examples, where \(a \in \{\mathrm{in}, \mathrm{mid}\}\).
We compute
\begin{equation}
g_{\ell+1} = M_{\ell+1}(Q_{\mathrm{mid}}) - M_{\ell+1}(Q_{\mathrm{in}}),
\end{equation}
and assign ES-In to layer \(\ell+1\) when \(g_{\ell+1} \leq \tau\), and ES-Mid otherwise.
Here, \(\tau\) is a tolerance threshold: if the advantage of ES-Mid over ES-In does not exceed \(\tau\), we use the earlier anchor for a longer preparation window; otherwise, we retain ES-Mid for higher fidelity.

The resulting layer-wise assignment remains fixed during inference.
This offline procedure does not require the target-layer query during routing, and preserves the underlying selector unchanged.
Section~\ref{sec:proxy_fidelity} analyzes how ES-In, ES-Mid, and ES-Mix behave across models and layer structures.

\subsection{ESelector Instantiations}
\label{sec:eselector_instantiations}

EarlySelectKV can wrap query-conditioned selectors whose routing decisions are driven by a query signal.
We instantiate it on Quest, Loki, and RocketKV, with RocketKV serving as the primary experimental setting.
RocketKV is built on Hybrid Sparse Attention (HSA), which partitions the historical KV cache into sequence pages and represents each page with lightweight summaries, such as \(K_{\max}\) and \(K_{\min}\), for query-conditioned page scoring.
For all instantiations, EarlySelectKV preserves the original routing view, budget, and scoring rule, and replaces only the routing query with the proxy query.
We evaluate the additional Quest and Loki instantiations in Section~\ref{sec:other_selector_results}.

\section{Experiments}

\subsection{Experimental Setup}

The main evaluation covers four long-context language models: Llama-3.1-8B-Instruct and Llama-3-8B-Instruct~\citep{llama3}, Mistral-7B-Instruct-v0.2~\citep{mistral7b}, and Qwen2.5-7B-Instruct~\citep{qwen2.5}. We use LongBench~\citep{LongBench}, Needle-in-a-Haystack (NIAH)~\citep{NIAH}, and RULER~\citep{RULER} to test general long-context understanding, long-range retrieval, and synthetic long-context reasoning.

Our main quality comparison is between RocketKV and ES-RocketKV. In these evaluations, they differ only in the routing-query source: RocketKV uses the target-layer real query, whereas ES-RocketKV uses a proxy query from a preceding hidden state. We also report FullCache as the dense-attention reference; it attends to the entire KV cache without sparse selection. Unless otherwise stated, EarlySelectKV uses the ES-Mid anchor for stronger proxy-query representations. For stability, all sparse methods keep layer 0 dense and apply sparse selection from layer 1 onward. Since ESelector precomputes indices before the target-layer keys are available, we reserve the local budget outside routing and keep the most recent tail page by default. ESelector routes only over the remaining historical cache.

\subsection{Benchmark Performance}
\label{sec:accuracy_results}

\begin{figure*}[t]
    \centering
    \includegraphics[width=\linewidth]{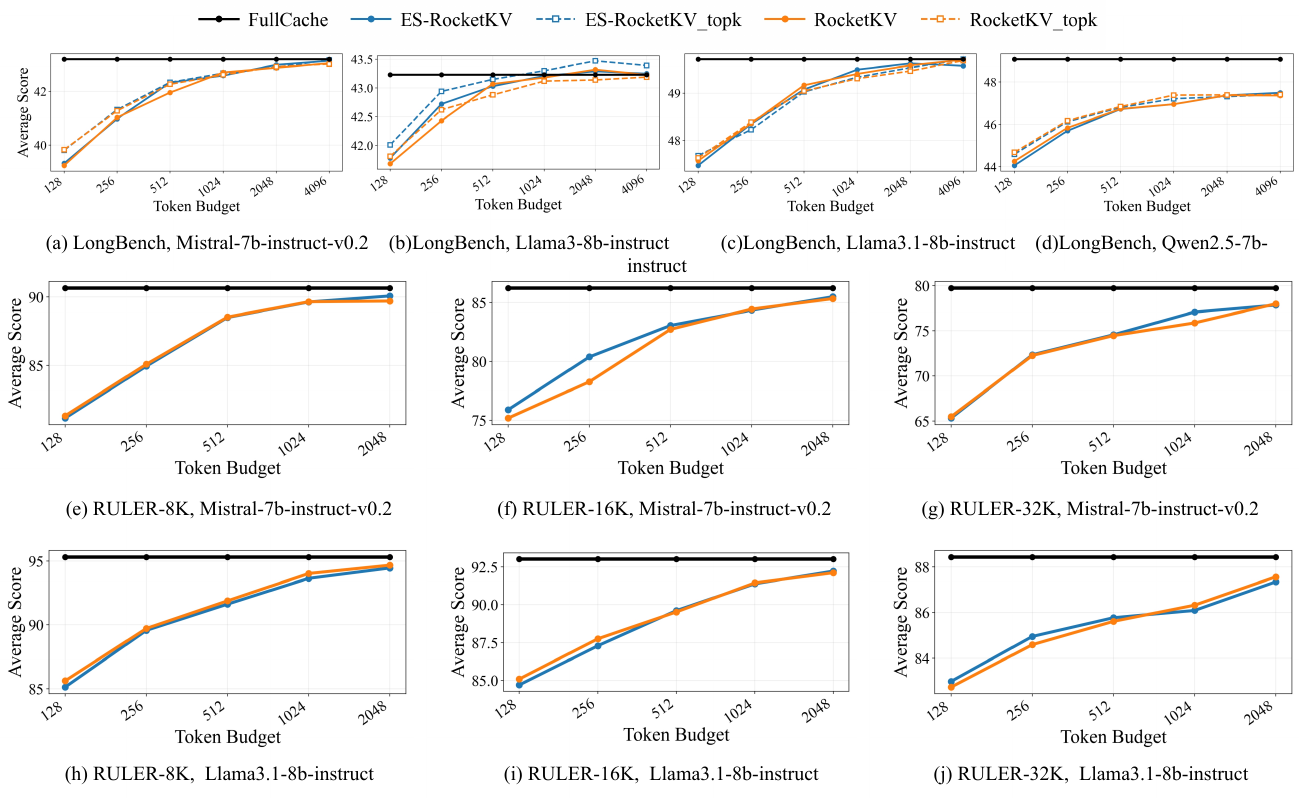}
    \caption{
    Task score comparison between ES-RocketKV and RocketKV on LongBench and RULER.
    The first row reports LongBench results, while the second and third rows report RULER results under different context lengths.
    }
    \label{fig:longbench_ruler}
\end{figure*}

Unless otherwise stated, the default quality sweep spans 128 to 4096 selected tokens; NIAH and anchor diagnostics also include a budget of 64.
We report average benchmark scores for LongBench and accuracy for NIAH and RULER.
Detailed per-task results are reported in Appendix~\ref{app:detailed_benchmark_results}.

\paragraph{LongBench Benchmark.}
Figure~\ref{fig:longbench_ruler} compares FullCache with two routing variants of RocketKV and ES-RocketKV. Solid lines use the default HSA routing, while dashed lines use exact Stage-2 Top-K routing. The figure reports averages; detailed subtask scores are in Appendix~\ref{app:detailed_benchmark_results}.

ES-RocketKV closely matches RocketKV across all evaluated settings. Since the two methods differ only in the routing query, this confirms that RocketKV's selection can be reliably approximated from an early proxy query. Averaged over LongBench, ES-RocketKV differs by 0.018 points; at the 128-token budget, by 0.03 points; across five backbones (7B--32B), every mean stays within 0.1 points, indicating no systematic loss in the averaged scores.

As the budget increases, the gap to FullCache narrows, indicating that the sparse budget, rather than proxy approximation, is the main source of degradation. Exact Stage-2 Top-K routing does not consistently outperform HSA routing, suggesting that the highest-attention tokens are not always the best predictor of task scores on generation-heavy tasks.

\paragraph{NIAH Benchmark.}
Table~\ref{tab:niah} reports representative NIAH passkey retrieval results; the budget sweep and detailed setup are provided in Appendix~\ref{app:niah_details}.
Across the representative settings in Table~\ref{tab:niah}, the mean ES-minus-RocketKV difference is only \(+0.25\) percentage points.
The results are identical on the Llama and Mistral models and remain comparable on the more budget-sensitive Qwen2.5-7B-Instruct. In these settings, proxy-query routing preserves the evidence-localization signal needed for long-context retrieval.

\begin{figure*}[t]
    \centering
    \includegraphics[width=\linewidth]{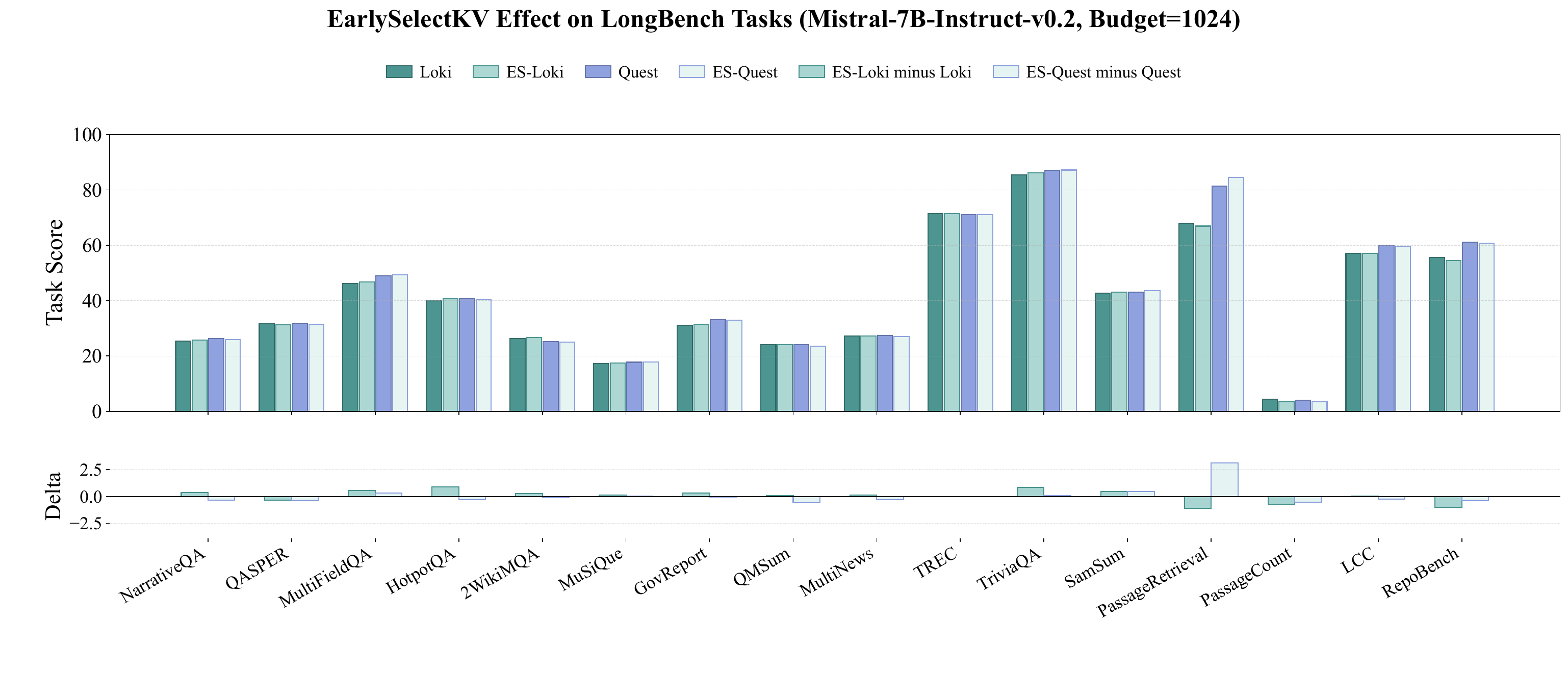}
    \caption{
    Additional selector results on LongBench using Mistral-7B-Instruct-v0.2 with a per-query KV-access budget of 1024 tokens.
    The upper panel reports task-level scores, and the lower panel shows the score difference between each ES variant and its corresponding base selector.
    }
    \label{fig:additional_selector_results}
\end{figure*}

\paragraph{RULER Benchmark.}
We evaluate ES-RocketKV on RULER-8K, 16K, and 32K (rows 2--3 of Figure~\ref{fig:longbench_ruler}); detailed scores are provided in Appendix~\ref{app:detailed_benchmark_results}. RULER covers retrieval, tracking, counting, and QA, providing a broad test of long-context sparse selection.

Across all settings, the mean ES-minus-RocketKV difference is approximately \(-0.01\) points.
ES-RocketKV also follows RocketKV's degradation as context length increases, indicating that early routing preserves the selector's accuracy--budget behavior.

\subsection{Applicability Beyond RocketKV}
\label{sec:other_selector_results}

To test whether EarlySelectKV is specific to RocketKV, we further apply the same selector-preserving transformation to Quest and Loki without changing their routing rules.
Quest uses query-conditioned page-level routing, while Loki relies on lightweight query-dependent scoring.

Figure~\ref{fig:additional_selector_results} reports task-level LongBench results on Mistral-7B-Instruct-v0.2 with a per-query KV-access budget of 1024. Table~\ref{tab:selector_budget_quality} in Appendix~\ref{app:longbench_details} extends this evaluation across additional selectors, budgets, and model scales.
Across the eight Quest/Loki comparisons on the two 7B backbones, the mean ES-minus-base difference is \(-0.07\) points. Together with the 1024-budget results, this supports the applicability of early proxy-query routing beyond RocketKV, while limiting the claim to the evaluated selector--budget pairs.

Unlike the inference-time selectors above, InfLLM-V2 is a trainable sparse-attention architecture that reuses dense-attention parameters and switches from dense attention on short inputs to sparse attention on long inputs.
We integrate EarlySelectKV into the released InfLLM-V2-Long-Sparse-Base checkpoint~\citep{InfLLMV2} without retraining.
Relative to the checkpoint's native real-query routing, EarlySelectKV with the exact \(W_Q\) changes the average across the four task scores by \(+0.26\) points, whereas the rank-256 approximation yields a change of \(-0.28\) points.
This setup tests integration after sparse-attention training. Incorporating EarlySelectKV during pretraining or long-context adaptation may improve proxy fidelity by allowing the hidden states, query projections, and routing behavior to co-adapt.
Appendix~\ref{app:trainable_sparse} provides detailed results and the evaluation protocol.

\begin{table}[t]
\centering
\small
\setlength{\tabcolsep}{8pt}
\renewcommand{\arraystretch}{1.08}
\begin{tabular}{c ccc}
\toprule
\multirow{2}{*}{Budget} & \multicolumn{3}{c}{Method} \\
\cmidrule(lr){2-4}
 & TopK & RocketKV & ES-RocketKV \\
\midrule
\rowcolor{gray!15}
\multicolumn{4}{c}{\textit{Mistral-7B-Instruct-v0.2}} \\
64  & 100.0 & \textbf{100.0} & \textbf{100.0} \\
128 & 100.0 & \textbf{100.0} & \textbf{100.0} \\
\midrule
\rowcolor{gray!15}
\multicolumn{4}{c}{\textit{Llama-3-8B-Instruct}} \\
64  & 100.0 & \textbf{98.0}  & \textbf{98.0} \\
128 & 100.0 & \textbf{100.0} & \textbf{100.0} \\
\midrule
\rowcolor{gray!15}
\multicolumn{4}{c}{\textit{Llama-3.1-8B-Instruct}} \\
64  & 100.0 & \textbf{88.0}  & \textbf{88.0} \\
128 & 100.0 & \textbf{100.0} & \textbf{100.0} \\
\midrule
\rowcolor{gray!15}
\multicolumn{4}{c}{\textit{Qwen2.5-7B-Instruct}} \\
64  & 76.0 & \textbf{48.0} & \textbf{48.0} \\
128 & 76.0 & 64.0 & \textbf{66.0} \\
\bottomrule
\end{tabular}
\caption{Matching accuracy (\%) on NIAH/Passkey under low per-query KV-access budgets. Across four instruction-tuned backbones, ES-RocketKV closely matches RocketKV in passkey retrieval accuracy.}
\label{tab:niah}
\end{table}

\begin{figure*}[t]
    \centering
    \includegraphics[width=\linewidth]{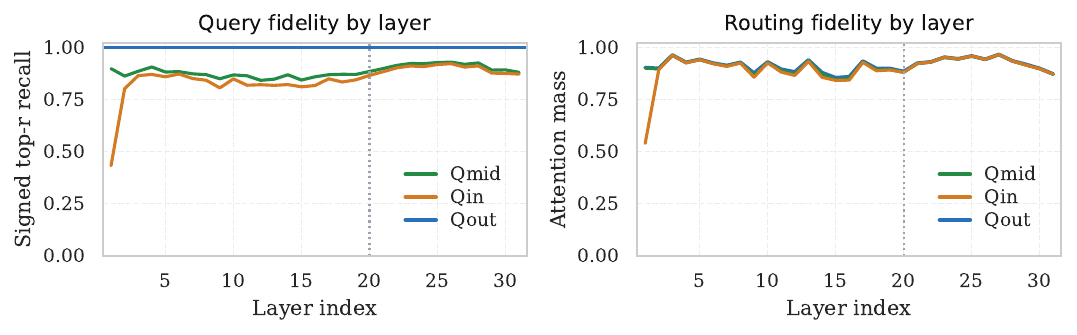}
    \caption{
    Proxy-query fidelity across the decoder layers of Mistral-7B-Instruct-v0.2.
    Left: signed top-\(r\) channel recall of \(Q_{\mathrm{in}}\) and \(Q_{\mathrm{mid}}\) with respect to \(Q_{\mathrm{out}}\).
    Right: attention mass retained by tokens selected using each query, evaluated under the \(Q_{\mathrm{out}}\) attention distribution.
    The dotted line marks the transition layer used by ES-Mix.
    }
    \label{fig:proxy_fidelity}
\end{figure*}

\subsection{Proxy Fidelity and Anchor Selection}
\label{sec:proxy_fidelity}

ES-RocketKV closely matches native RocketKV at the task level. We examine whether its proxy queries preserve the routing evidence used by the native selector.
For next layer \(\ell+1\), let \(Q_{\mathrm{out}}\) denote its native routing query, and let \(Q_{\mathrm{in}}\) and \(Q_{\mathrm{mid}}\) denote proxy queries constructed from \(H_{\ell}^{\mathrm{in}}\) and \(H_{\ell}^{\mathrm{mid}}\), respectively.
Because EarlySelectKV changes only the routing query, we evaluate proxy fidelity through routing behavior rather than exact query reconstruction.

We use two diagnostics: signed top-\(r\) channel recall, which measures agreement on the dominant signed pre-RoPE query dimensions of \(Q_{\mathrm{out}}\), and retained attention mass, which measures how much \(Q_{\mathrm{out}}\)-based attention mass is covered by proxy-selected tokens.
The diagnostic setup is described in Appendix~\ref{app:proxy_diagnostics}.

Figure~\ref{fig:proxy_fidelity} shows that \(Q_{\mathrm{mid}}\) provides higher signed-channel recall and retained attention mass in early and middle layers, while the two anchors converge in later layers.
This pattern motivates ES-Mix, which uses ES-Mid for anchor-sensitive layers and ES-In where their routing fidelity is comparable.
Appendix~\ref{app:mix_stability} confirms stable profiles across LongAlpaca-12k, Qasper, and QMSum, with pairwise layer-gap correlations of at least \(0.9724\) and assignment agreement of at least \(96.8\%\).
Across 21 LongBench and 30 RULER settings, ES-Mix differs from ES-Mid by \(-0.006\) points on average, indicating that it extends the preparation window while preserving, rather than improving, task quality.
Detailed comparisons appear in Appendix~\ref{app:anchor_quality}.

\subsection{Long-Generation Evaluation}
\label{sec:long_generation}

Autoregressive feedback can obscure whether routing mismatch itself grows with decode position.
To isolate this position effect, we compare RocketKV and ES-RocketKV on the same 20 MultiNews trajectories over decode positions 0 to 512.

\begin{figure*}[t]
    \centering
    \includegraphics[width=0.95\linewidth]{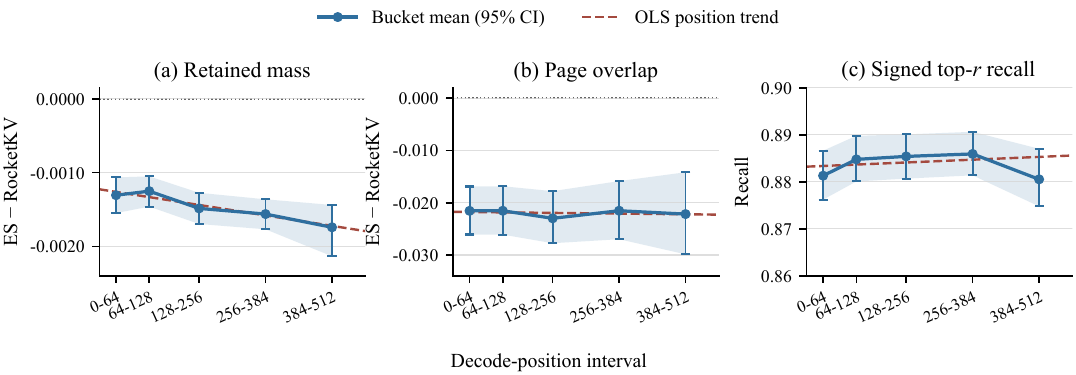}
    \caption{
    Fixed-trajectory routing fidelity over decode position on MultiNews.
    Points connect five interval means; shaded regions and vertical bars denote 95\% bootstrap confidence intervals, and dashed lines show OLS position trends.
    Retained mass and page overlap are ES-RocketKV-minus-RocketKV gaps, whereas signed top-\(r\) recall is measured between the proxy and target-layer queries.
    }
    \label{fig:long_generation_trajectory}
\end{figure*}

Across the five intervals in Figure~\ref{fig:long_generation_trajectory}, the retained-mass gap remains below \(0.002\) in magnitude.
Its small negative trend is not accompanied by a decline in page overlap or signed top-\(r\) recall, whose trend intervals both include zero.
The three metrics therefore do not show a consistent loss of routing fidelity as the shared trajectory lengthens.
Because this control holds model states fixed, it cannot measure feedback after their trajectories diverge.
The independent MATH-500 evaluation in Appendix~\ref{app:long_generation} allows each method to follow its own trajectory.
Across both tested models, ES-RocketKV remains within 1.0 percentage point of RocketKV, showing that early routing maintains comparable mathematical-reasoning accuracy under independent autoregressive generation at the evaluated budget.

\subsection{Runtime Benefits}
\label{sec:runtime}

\paragraph{Deployment Setup.}
We benchmark RocketKV and ES-RocketKV with Llama-3.1-8B-Instruct on a single NVIDIA RTX 6000D GPU using the TensorRT-LLM~\citep{TRT} PyTorch backend with CUDA Graphs enabled. Both methods use the same sparse-attention configuration and FP8 K-summary caches.

All runs use a Stage-1 prompt budget of 4096 tokens. The default Stage-2 configuration uses page size \(p=4\) and \(k=32\) selected pages, yielding \(pk=128\) historical tokens per query, with the local tail page reserved separately.

ES-RocketKV offloads proxy-query construction, routing, and K-summary state to an Intel Xeon Platinum CPU, using a BF16 rank-128 projection with AVX-512 BF16 dot products and FP32 accumulation. The full KV cache remains GPU-resident without page prefetch.
Because ES-RocketKV changes both when and where routing indices are produced, runtime results reflect deployment-level gains from anticipatory index availability under CPU-offloaded routing, rather than isolated proxy-query microbenchmarks.

\begin{figure}[t]
    \centering
    \includegraphics[width=\linewidth]{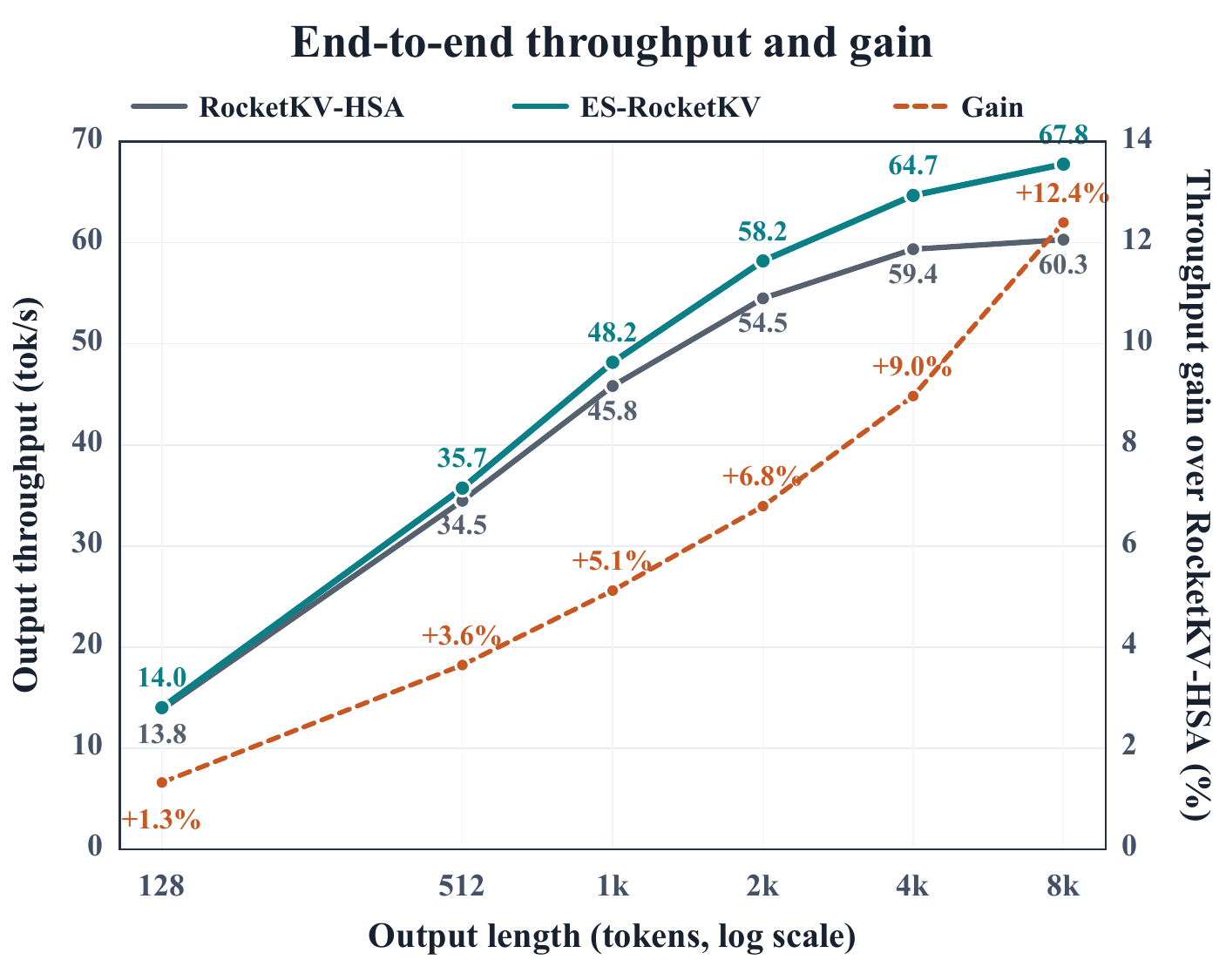}
    \caption{
    End-to-end output throughput on a 40K-token input with a 4096-token Stage-1 prompt budget, a 128-token Stage-2 selected-token budget, and batch size 1.
    Solid lines report the output throughput of RocketKV and ES-RocketKV, while the dashed line shows the relative throughput gain of ES-RocketKV over RocketKV.
    }
    \label{fig:throughput}
\end{figure}

\paragraph{Throughput Gain.}
Figure~\ref{fig:throughput} reports end-to-end output throughput on a 40K-token input with a 4K Stage-1 prompt budget and the default Stage-2 budget of 128 tokens. ES-RocketKV consistently improves throughput over RocketKV, with gains increasing from 1.3\% at 128 output tokens to 12.4\% at 8192 output tokens. Prefill dominates short-output requests, whereas longer outputs include more decode steps over which precomputed indices can reduce routing-related stalls on the GPU attention path.

Appendix~\ref{app:runtime_breakdown} separates post-context decoding from batch scaling. Decode-only gains rise from 7.67\% to 13.14\%; with 8192 output tokens, end-to-end gains remain between 10.5\% and 12.1\% across batch sizes 1 to 8. By contrast, the short-output benefit approaches zero at batch 8, showing that longer decodes better amortize fixed routing and synchronization costs.

\paragraph{GPU Memory Savings.}
Offloading routing also removes the need to keep K-summary auxiliary states on the GPU.
In our configuration, the allocation-derived saving is about 5.3 GiB, corresponding to 11.1\% of the KV and K-summary pool.
This memory reduction reflects the CPU-offloaded deployment choice.
EarlySelectKV makes this placement practical by exposing selected indices early enough to stage routing and metadata preparation before the target attention layer.

\section{Conclusion}
\label{sec:conclusion}

We presented \textsc{EarlySelectKV}, a training-free transformation for query-conditioned sparse KV selectors.
By replacing the target-layer routing query with a proxy from an earlier hidden state, EarlySelectKV makes selected KV indices available before the target attention layer while preserving the selector structure.
Instantiated primarily on RocketKV as ES-RocketKV, EarlySelectKV closely matches native RocketKV on LongBench, Needle-in-a-Haystack, and RULER in task score and selection behavior.
Proxy anchors reveal a trade-off between preparation window and routing stability. Runtime results show that early index availability can improve throughput in the evaluated deployment and CPU-offloaded routing can reduce GPU-resident auxiliary state. EarlySelectKV highlights routing timing as an additional optimization dimension for sparse long-context decoding.

\section*{Limitations}

\textsc{EarlySelectKV} changes routing timing rather than selector complexity.
Its runtime benefit depends on how much proxy construction, routing, and metadata preparation fit within the available preparation window.
Gains may be smaller for short outputs, fine-grained or expensive routers, large batches, and resource-contended execution.

Our runtime study covers a single-GPU TensorRT-LLM deployment with CPU-offloaded routing and GPU-resident KV states.
It does not cover multi-GPU serving, KV-page prefetch or offloading, or larger-model runtime scaling, so the measured throughput and memory gains may vary across systems and workloads.

ES-Mix uses a fixed offline layer profile and may require re-profiling under substantial domain shifts.
Most quality experiments focus on RocketKV; the InfLLM-V2 analysis covers one checkpoint without runtime measurements, and the long-generation evaluation is limited to the tested models and horizons.

\section*{Acknowledgments}

This work was supported by the National Key Research and Development Program of China (No.~2024YFE0210600), the National Natural Science Foundation of China (No.~U24A20236), the CAS Project for Young Scientists in Basic Research (Nos.~YSBR-090 and YSBR-118), and the Beijing Natural Science Foundation (No.~L2608015).

Generative AI assistance was used solely for language polishing, proofreading, and submission-related formatting. The authors take full responsibility for the manuscript.

\bibliography{custom}

\appendix

\section{Effect of Low-Rank Proxy Projection}
\label{app:rank_tradeoff}

We study how the approximation rank affects low-rank proxy-query construction.
Given a rank-\(r\) approximation of the target-layer query projection,
\begin{equation}
    W_q^{\ell+1} \approx U_r^{\ell+1} \Sigma_r^{\ell+1} (V_r^{\ell+1})^\top,
\end{equation}
the proxy query is computed as
\begin{equation}
    q_{t,\mathrm{proxy}}^{\ell+1}
    =
    U_r^{\ell+1} \Sigma_r^{\ell+1} (V_r^{\ell+1})^\top
    \operatorname{LN}_{\ell+1}(\tilde{H}_{t,\ell}).
\end{equation}
The rank \(r\) controls a direct trade-off between cost and fidelity: smaller ranks reduce proxy-query construction cost, while larger ranks better preserve the target-layer query space.

\begin{figure*}[t]
    \centering
    \includegraphics[width=\linewidth]{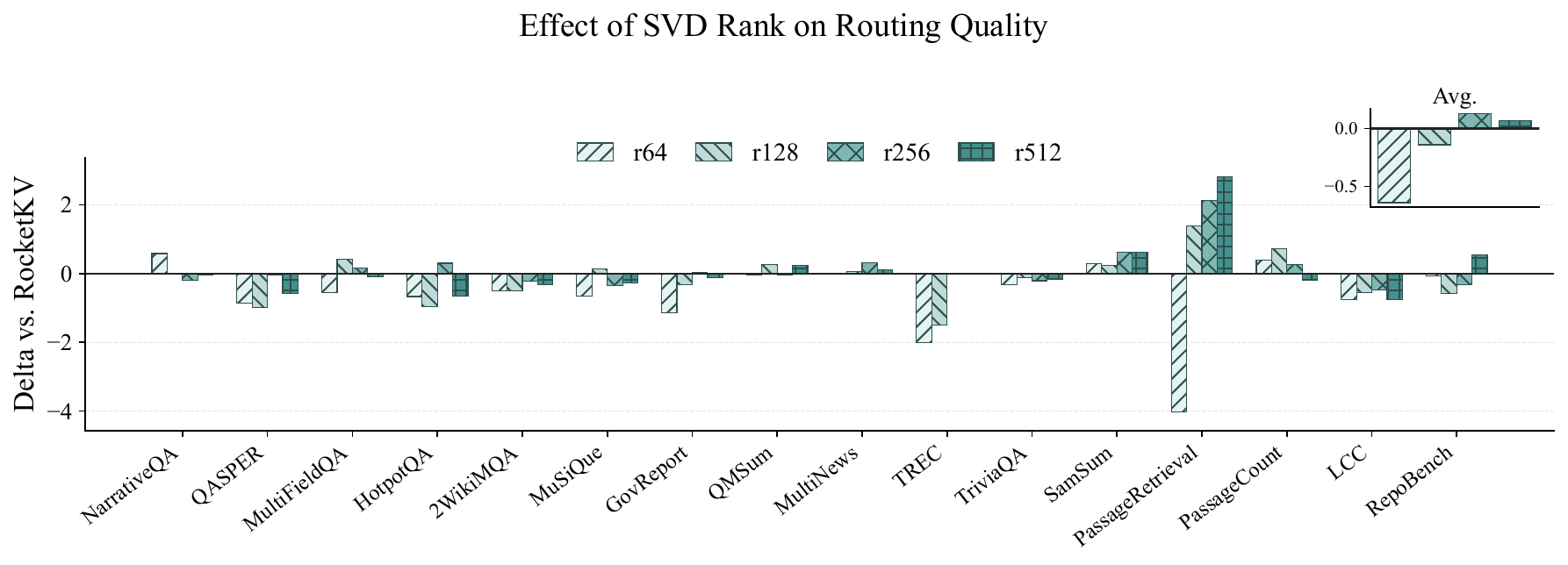}
    \caption{
    Effect of the low-rank proxy-query projection on LongBench using Mistral-7B-Instruct-v0.2 with a per-query KV-access budget of 512.
    Scores are compared against RocketKV with the full target-layer query projection.
    }
    \label{fig:rank_tradeoff}
\end{figure*}

Figure~\ref{fig:rank_tradeoff} reports LongBench results on Mistral-7B-Instruct-v0.2 with a per-query KV-access budget of 512.
The original RocketKV obtains an average score of \(41.96\).
A low rank of \(r=64\) reduces the average score to \(41.32\), mainly due to larger drops on retrieval-sensitive tasks such as PassageRetrieval.
Increasing the rank quickly closes the gap: \(r=128\) reaches \(41.82\), while \(r=256\) and \(r=512\) obtain \(42.09\) and \(42.03\), respectively. Performance saturates around \(r=256\), with little benefit from a further increase. We use rank-128 in the CPU-routing experiments to reduce host-side routing cost; the higher-rank results quantify the available trade-off between fidelity and cost when more proxy-query computation is acceptable.

\section{Detailed Benchmark Results}
\label{app:detailed_benchmark_results}

We report the numerical results omitted from the main text: LongBench task-level and average scores, including larger-model and additional-selector results; the complete NIAH budget sweep; and RULER averages.

\subsection{LongBench Results}
\label{app:longbench_details}

\paragraph{Task-Level Results.}
\Cref{tab:longbench_task_llama3_8b_instruct,tab:longbench_task_llama3_1_8b_instruct,tab:longbench_task_mistral_7b_instruct_v0_2,tab:longbench_task_qwen2_5_7b_instruct} report the full LongBench subtask scores for each evaluated model.

\paragraph{Average Results.}
Table~\ref{tab:longbench_avg} reports averaged LongBench scores under different models and KV-access budgets.
The values underlie the LongBench trends in the first row of Figure~\ref{fig:longbench_ruler}.

\begin{table}[t]
\centering
\small
\setlength{\tabcolsep}{5pt}
\renewcommand{\arraystretch}{1.08}
\begin{tabular}{c cccc}
\toprule
\multirow{2}{*}{Budget} & \multicolumn{4}{c}{Method} \\
\cmidrule(lr){2-5}
 & FullCache & TopK & RocketKV & ES-RocketKV \\
\midrule
\rowcolor{gray!15}
\multicolumn{5}{c}{\textit{Llama-3-8B-Instruct}} \\
128  & 43.23 & 43.72 & 41.68 & \textbf{41.78} \\
256  & 43.23 & 43.58 & 42.43 & \textbf{42.72} \\
512  & 43.23 & 43.50 & \textbf{43.07} & 43.03 \\
1024 & 43.23 & 43.31 & 43.18 & \textbf{43.21} \\
2048 & 43.23 & 43.31 & \textbf{43.32} & 43.29 \\
4096 & 43.23 & 43.36 & 43.23 & \textbf{43.25} \\
\midrule
\rowcolor{gray!15}
\multicolumn{5}{c}{\textit{Llama-3.1-8B-Instruct}} \\
128  & 49.72 & 49.66 & \textbf{47.57} & 47.47 \\
256  & 49.72 & 49.72 & \textbf{48.36} & 48.35 \\
512  & 49.72 & 49.66 & \textbf{49.17} & 49.08 \\
1024 & 49.72 & 49.70 & 49.41 & \textbf{49.50} \\
2048 & 49.72 & 49.56 & 49.59 & \textbf{49.63} \\
4096 & 49.72 & 49.73 & \textbf{49.71} & 49.58 \\
\midrule
\rowcolor{gray!15}
\multicolumn{5}{c}{\textit{Mistral-7B-Instruct-v0.2}} \\
128  & 43.20 & 42.02 & 39.24 & \textbf{39.32} \\
256  & 43.20 & 42.68 & \textbf{41.03} & 40.98 \\
512  & 43.20 & 43.01 & 41.96 & \textbf{42.33} \\
1024 & 43.20 & 43.04 & \textbf{42.71} & 42.59 \\
2048 & 43.20 & 43.02 & 42.88 & \textbf{42.99} \\
4096 & 43.20 & 43.05 & 43.05 & \textbf{43.14} \\
\midrule
\rowcolor{gray!15}
\multicolumn{5}{c}{\textit{Qwen2.5-7B-Instruct}} \\
128  & 49.07 & 47.34 & \textbf{44.26} & 44.06 \\
256  & 49.07 & 47.36 & \textbf{45.84} & 45.70 \\
512  & 49.07 & 47.66 & \textbf{46.73} & 46.72 \\
1024 & 49.07 & 47.46 & \textbf{46.95} & \textbf{46.95} \\
2048 & 49.07 & 47.50 & \textbf{47.37} & \textbf{47.37} \\
4096 & 49.07 & 47.60 & 47.36 & \textbf{47.49} \\
\bottomrule
\end{tabular}
\caption{Average LongBench scores of different methods under different per-query KV-access budgets.}
\label{tab:longbench_avg}
\end{table}

\paragraph{Larger-Model and Selector Generalization.}
Table~\ref{tab:selector_budget_quality} reports LongBench averages on Qwen2.5-32B-Instruct together with additional Quest and Loki comparisons on Qwen2.5-7B and Mistral-7B.
For RocketKV, the ES-minus-base difference averages \(+0.015\) points across budgets 256 and 1024; for Quest at budget 256, the difference is \(-0.01\) points.
Across these settings, the ES variants closely match their base selectors.

\begin{table}[t]
\centering
\footnotesize
\setlength{\tabcolsep}{3pt}
\renewcommand{\arraystretch}{1.08}
\begin{tabular}{@{}llrrrr@{}}
\toprule
Model & Selector & Budget & Base & +ES & $\Delta$ \\
\midrule
\multirow{3}{*}{Qwen2.5-32B}
 & Quest    & 256  & 47.62 & 47.61 & $-0.01$ \\
 & RocketKV & 256  & 47.10 & 47.14 & $\mathbf{+0.04}$ \\
 & RocketKV & 1024 & 48.16 & 48.15 & $-0.01$ \\
\midrule
\multirow{2}{*}{Qwen2.5-7B}
 & Loki  & 128  & 38.36 & 38.44 & $\mathbf{+0.08}$ \\
 & Quest & 128  & 43.30 & 42.93 & $-0.37$ \\
\midrule
\multirow{6}{*}{Mistral-7B}
 & Loki  & 128  & 29.51 & 28.65 & $-0.86$ \\
 & Quest & 128  & 39.29 & 39.11 & $-0.18$ \\
 & Loki  & 2048 & 41.78 & 42.08 & $\mathbf{+0.30}$ \\
 & Quest & 2048 & 42.84 & 42.95 & $\mathbf{+0.11}$ \\
 & Loki  & 4096 & 42.54 & 42.88 & $\mathbf{+0.34}$ \\
 & Quest & 4096 & 43.09 & 43.13 & $\mathbf{+0.04}$ \\
\bottomrule
\end{tabular}
\caption{Additional LongBench averages across model scales, selectors, and KV-access budgets. \(\Delta\) denotes ES minus the corresponding base selector.}
\label{tab:selector_budget_quality}
\end{table}

\subsection{Needle-in-a-Haystack Results}
\label{app:niah_details}

\begin{figure*}[!t]
    \centering
    \begin{subfigure}[t]{0.32\linewidth}
        \centering
        \includegraphics[width=\linewidth]{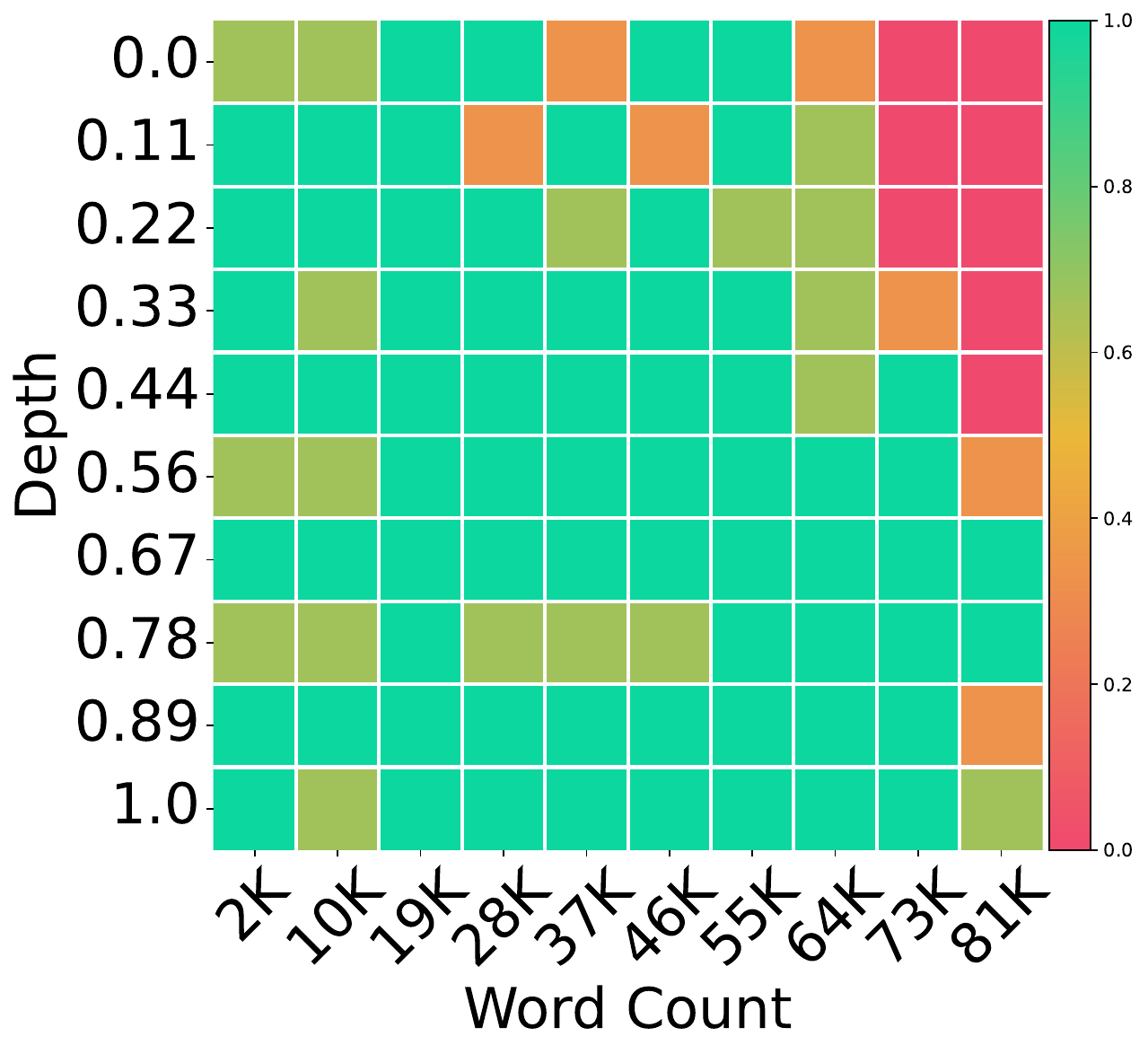}
        \caption{RocketKV}
        \label{fig:qwen_heatmap_rocketkv}
    \end{subfigure}
    \hfill
    \begin{subfigure}[t]{0.32\linewidth}
        \centering
        \includegraphics[width=\linewidth]{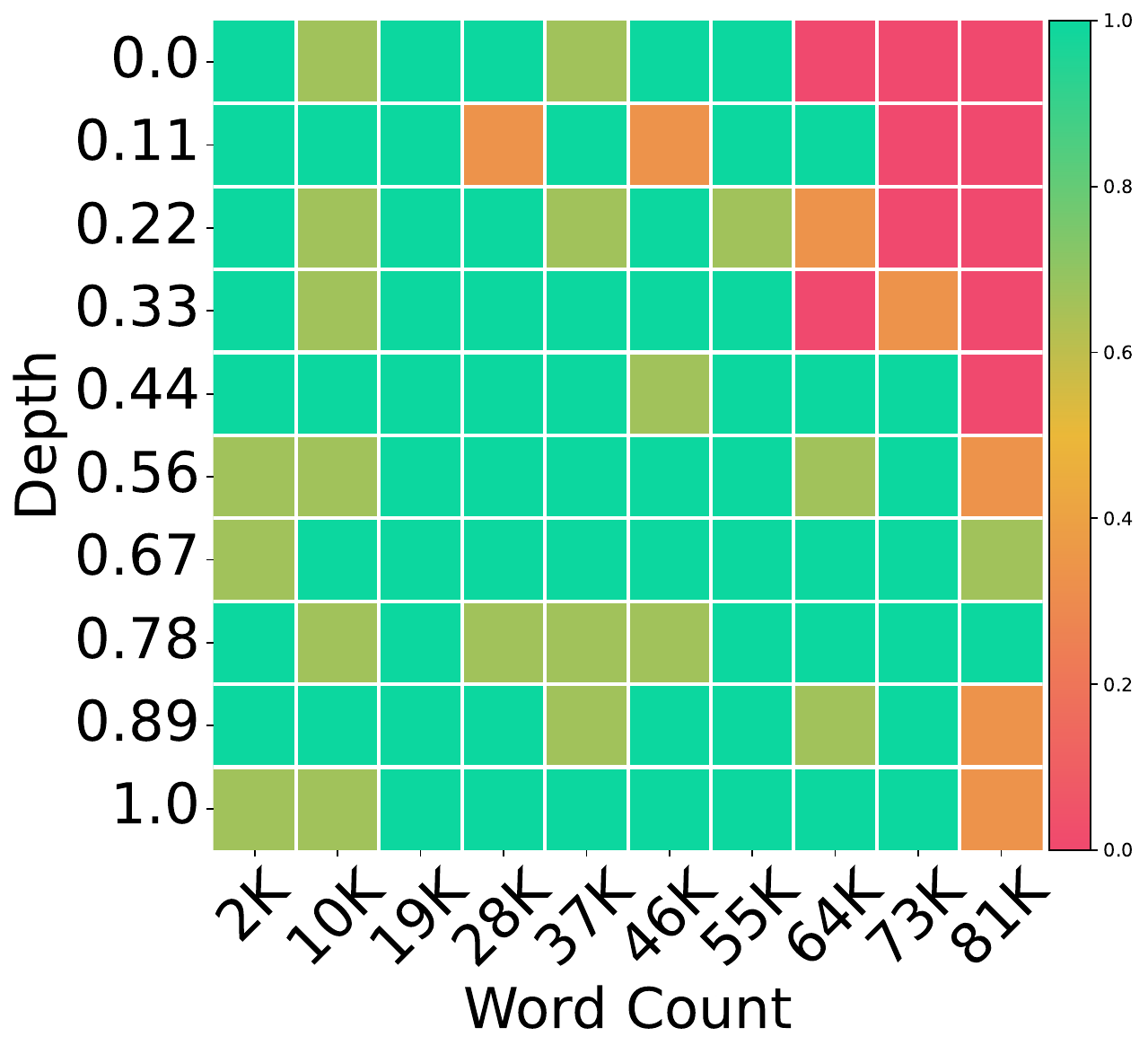}
        \caption{ES-In}
        \label{fig:qwen_heatmap_es_in}
    \end{subfigure}
    \hfill
    \begin{subfigure}[t]{0.32\linewidth}
        \centering
        \includegraphics[width=\linewidth]{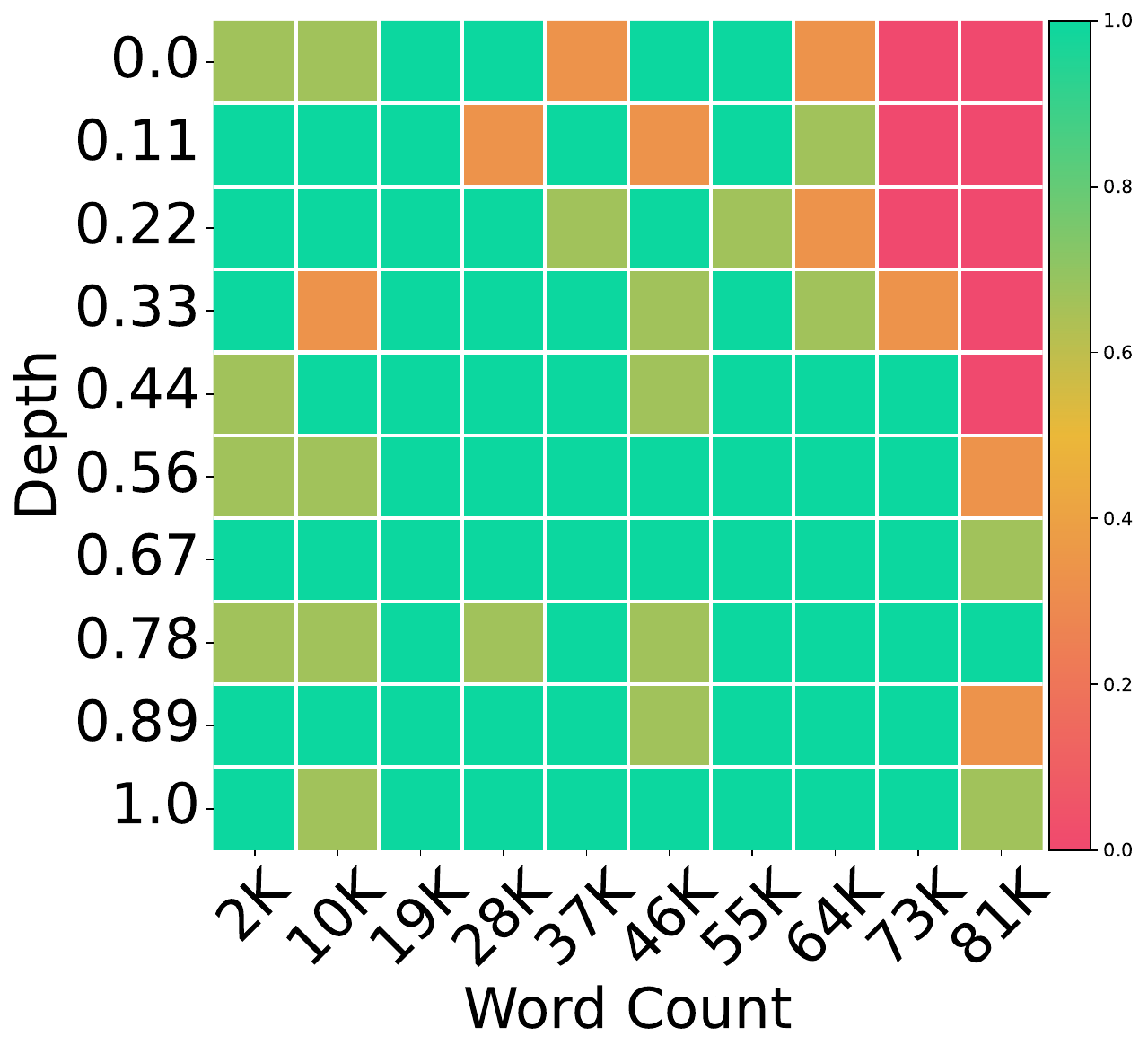}
        \caption{ES-Mid}
        \label{fig:qwen_heatmap_es_mid}
    \end{subfigure}
    \caption{
    NIAH/Passkey exact-match accuracy heatmaps on Qwen2.5-7B-Instruct across context lengths and needle depths. All methods use a per-query KV-access budget of 256 tokens.
    }
    \label{fig:qwen_heatmap_compare}
\end{figure*}

We summarize the Needle-in-a-Haystack (NIAH)/Passkey budget sweep omitted from the main text.
Accuracy is measured by exact match on the inserted passkey.
For Qwen2.5-7B-Instruct, we sweep per-query KV-access budgets from 64 to 4096 and compare RocketKV, the available ES-RocketKV variants, and Exact-TopK.
For the other models, Table~\ref{tab:niah_full} reports the available low-budget evaluations and compact threshold summaries for settings that reach perfect retrieval.
Figure~\ref{fig:qwen_heatmap_compare} compares NIAH/Passkey exact-match accuracy for RocketKV, ES-In, and ES-Mid across context lengths and needle depths.

\subsection{RULER Results}
\label{app:ruler_details}

Table~\ref{tab:ruler_average} reports averaged RULER scores under different context lengths and KV-access budgets.
The values underlie the RULER trends in the second and third rows of Figure~\ref{fig:longbench_ruler}.

\section{Proxy Fidelity and Anchor Selection}

\subsection{Layer-Wise Proxy Diagnostics}
\label{app:proxy_diagnostics}

\begin{figure*}[t]
    \centering
    \includegraphics[width=\linewidth]{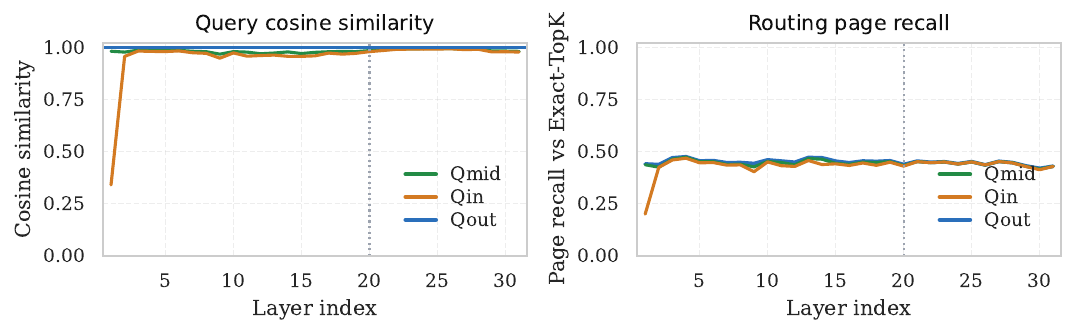}
    \caption{
    Auxiliary proxy-query diagnostics on the LongAlpaca-12k calibration set.
    Left: cosine similarity to \(Q_{\mathrm{out}}\).
    Right: page recall against Exact-TopK selection.
    }
    \label{fig:appendix_proxy_fidelity}
\end{figure*}

Using Mistral-7B-Instruct-v0.2 and a KV-access budget of 1024, we collect 38,967 decode-layer records from 10 LongAlpaca-12k prompts.
For each pre-RoPE query-head vector \(q\in\mathbb{R}^{d}\), let
\begin{equation}
    S_r(q)=\{(j,\operatorname{sign}(q_j)):j\in\operatorname{TopK}(|q|,r)\}.
\end{equation}
We define signed top-\(r\) recall between a proxy query \(q_p\) and the real query \(q_r\) as
\begin{equation}
    \operatorname{Recall}_r(q_p,q_r)
    =\frac{|S_r(q_p)\cap S_r(q_r)|}{r}.
\end{equation}
Here, \(r=\operatorname{clip}(\operatorname{round}(dp/c_2),1,d)\) is the per-example HSA channel budget, where \(d\) is the head dimension (128 in these experiments), \(p\) is the HSA chunk size, and \(c_2\) is the Stage-2 compression ratio. We first average recall over query heads within each layer/decode record. Figure~\ref{fig:proxy_fidelity} then averages over decoded records and examples at each layer, while Figure~\ref{fig:long_generation_trajectory} first averages over routed layers and decode steps within each example and interval and then averages across examples.
Figure~\ref{fig:appendix_proxy_fidelity} supplements the main-text diagnostics with cosine similarity and Exact-TopK page recall.
All four metrics follow the same layer-wise trend: the retained-mass advantage of \(Q_{\mathrm{mid}}\) over \(Q_{\mathrm{in}}\) is \(0.0220\) before layer 20 and \(0.0003\) afterward.
This convergence motivates ES-Mix: it uses \(Q_{\mathrm{mid}}\) where the anchor affects routing and switches to the earlier \(Q_{\mathrm{in}}\) once the two proxies become comparable.

\subsection{Cross-Domain Profile Stability}
\label{app:mix_stability}
We test whether ES-Mix produces similar layer profiles when calibrated on different domains. For each of LongAlpaca-12k, Qasper, and QMSum, we select 50 prompts with seed 42 and reuse the same subset across models.
All profiles use a KV-access budget of 1024 and \(\tau=0.01\). Each backbone has 31 routed layers because layer 0 remains dense.
For every pair of domains, we compare the layer-wise fidelity gaps using Pearson correlation and measure how often the resulting anchor assignments agree. Table~\ref{tab:esmix_domain_stability} reports the range across the three domain pairs.

\begin{table}[t]
\centering
\small
\setlength{\tabcolsep}{4pt}
\renewcommand{\arraystretch}{1.08}
\begin{tabular}{@{}lrrr@{}}
\toprule
Model & Routed & Pearson \(r\) & Agreement \\
\midrule
Mistral-7B   & 31 & 0.9988--0.9995 & 100.0\% \\
Llama-3-8B   & 31 & 0.9786--0.9975 & 96.8\%--100.0\% \\
Llama-3.1-8B & 31 & 0.9724--0.9942 & 100.0\% \\
\bottomrule
\end{tabular}
\caption{Cross-domain stability of ES-Mix profiles. Pearson correlations and assignment-agreement ranges are computed over the three pairwise comparisons among LongAlpaca-12k, Qasper, and QMSum.}
\label{tab:esmix_domain_stability}
\end{table}

The anchor assignments are identical across domains for Mistral and Llama-3.1. Llama-3 differs at only one of the 31 routed layers in the least consistent domain pair.
For Qwen2.5-7B-Instruct, the layers satisfying \(g_{\ell+1}\leq\tau\) at \(\tau=0.01\) do not form a contiguous late-layer ES-In region. Since the evaluated ES-Mix configuration uses a single ES-Mid-to-ES-In transition, we use ES-Mid for all routed layers and do not report ES-Mix results for this model.

\subsection{Anchor-Setting Quality}
\label{app:anchor_quality}
\Cref{tab:ruler_lookahead_variants,tab:longbench_lookahead_variants} report averaged LongBench and RULER scores for ES-In, ES-Mid, and ES-Mix.
The mean ES-Mix-minus-ES-Mid differences are \(+0.014\) points across 21 LongBench settings and \(-0.021\) points across 30 RULER settings, giving a combined mean of \(-0.006\) points.
The small, opposite-signed differences indicate that the mixed profile preserves ES-Mid-level quality across the evaluated settings.

\section{Long-Generation Diagnostics}
\label{app:long_generation}

\paragraph{Independent Generation.}
We evaluate the full test split of MATH-500 from \texttt{HuggingFaceH4/MATH-500}.\footnote{\url{https://huggingface.co/datasets/HuggingFaceH4/MATH-500}}
Each method generates its own trajectory; RocketKV and ES-RocketKV use a KV-access budget of 1024.
We use greedy decoding with at most 4096 new tokens and compare the extracted final answers after normalization.
Table~\ref{tab:math500_long_generation} reports the resulting accuracy.

\begin{table}[t]
\centering
\small
\setlength{\tabcolsep}{4pt}
\renewcommand{\arraystretch}{1.08}
\begin{tabular}{@{}lrrrr@{}}
\toprule
Model & Full & RocketKV & +ES & $\Delta$ \\
\midrule
Llama-3.1-8B & 45.6 & 44.4 & 45.4 & $+1.0$ \\
Qwen2.5-7B  & 72.4 & 74.2 & 73.4 & $-0.8$ \\
\bottomrule
\end{tabular}
\caption{Independent greedy-generation accuracy (\%) on MATH-500. RocketKV and ES-RocketKV use a per-query KV-access budget of 1024; \(\Delta\) denotes ES-RocketKV minus RocketKV.}
\label{tab:math500_long_generation}
\end{table}

\paragraph{Fixed-Trajectory Analysis.}
For the diagnostic in Section~\ref{sec:long_generation}, we use the first 20 MultiNews examples in LongBench order with Mistral-7B-Instruct-v0.2, a KV-access budget of 1024, and a maximum of 512 generated tokens.
The five position intervals \([0,64)\), \([64,128)\), \([128,256)\), \([256,384)\), and \([384,512)\) contain 20, 20, 20, 18, and 6 examples, respectively.
We fit OLS trends to the 84 example-interval observations and obtain 95\% confidence intervals from 5,000 example-level bootstrap resamples.

\section{Integration with InfLLM-V2}
\label{app:trainable_sparse}

We integrate EarlySelectKV into \texttt{InfLLM-V2} without retraining and evaluate the first 16 LongBench examples per task. Both variants use \(Q_{\mathrm{mid}}\) and differ only in whether \(W_Q\) is exact or approximated at rank 256.
\begin{table}[t]
\centering
\small
\setlength{\tabcolsep}{3.5pt}
\renewcommand{\arraystretch}{1.08}
\begin{tabular}{@{}lcrrr@{}}
\toprule
Task & Sparse & Real-\(Q\) & Exact-\(W_Q\) & Rank-256 \\
\midrule
GovReport & 13/16 & 25.21 & \textbf{25.91} & 24.68 \\
QMSum     & 15/16 & 24.05 & \textbf{24.47} & 23.80 \\
MuSiQue   & 16/16 &  \textbf{6.96} &  6.89 &  6.78 \\
HotpotQA  & 15/16 & \textbf{15.11} & \textbf{15.11} & 14.96 \\
\bottomrule
\end{tabular}
\caption{Post-hoc EarlySelectKV results on InfLLM-V2-Long-Sparse-Base. Real-\(Q\) is native routing; Exact-\(W_Q\) and Rank-256 use exact and rank-256 proxy projections. Scores cover all 16 examples per task; ``Sparse'' counts those activating sparse routing. GovReport/QMSum use ROUGE-L; MuSiQue/HotpotQA use QA F1.}
\label{tab:infllmv2_compatibility}
\end{table}

On the sparse-triggered subsets, the four-task average differences from native routing are \(+0.31\) points with the exact \(W_Q\) and \(-0.31\) points with the rank-256 approximation.

\section{Runtime Scope}
\label{app:runtime_breakdown}

\begin{figure}[t]
    \centering
    \includegraphics[width=\linewidth]{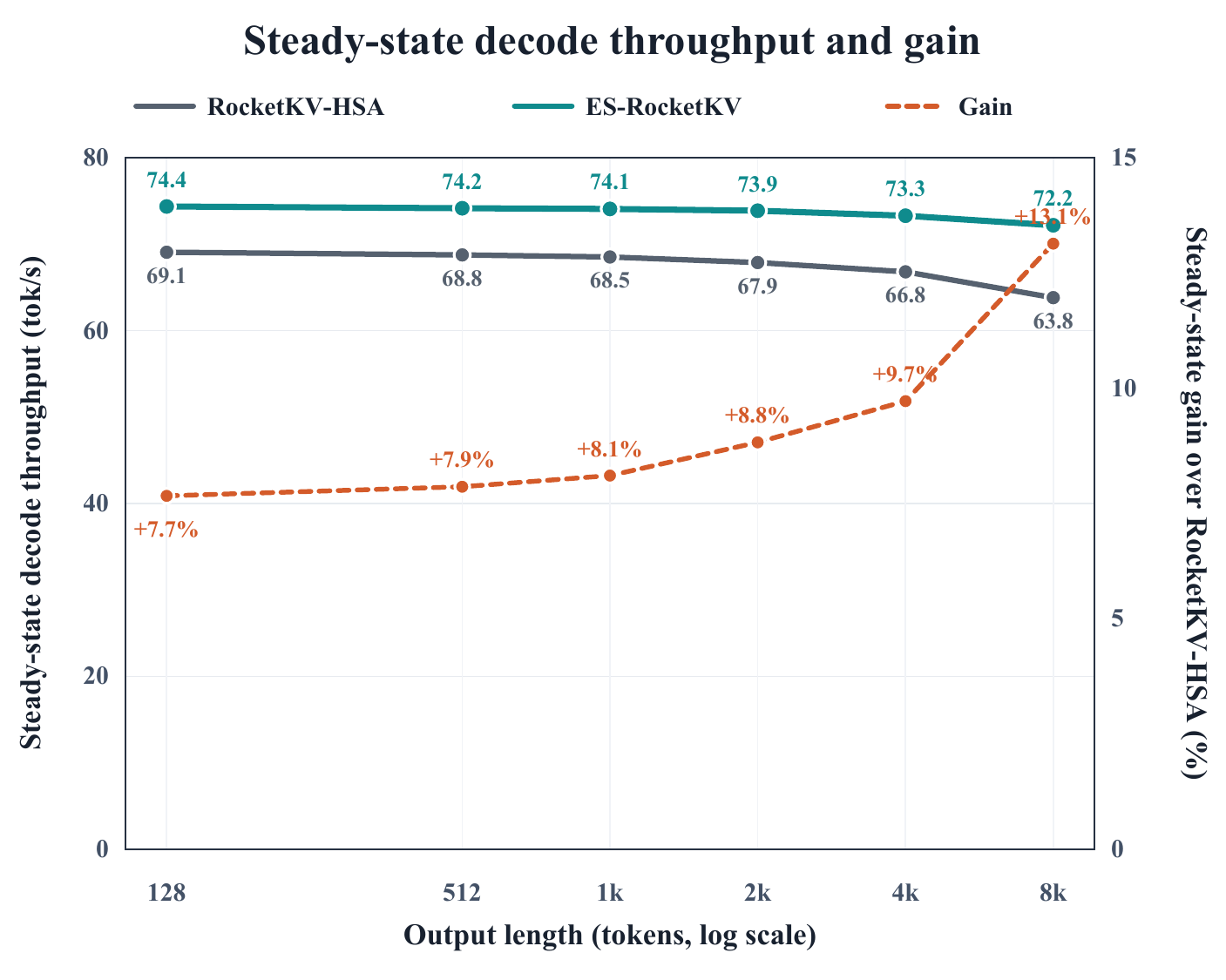}
    \caption{
    Steady-state decode throughput under different output lengths.
    The input length is 40K tokens, the Stage-1 prompt budget is 4096 tokens, and the Stage-2 selected-token budget is 128.
    Solid lines report post-context decode throughput of RocketKV-HSA and ES-RocketKV, while the dashed line reports the relative gain of ES-RocketKV.
    }
    \label{fig:decode_breakdown}
\end{figure}

Figure~\ref{fig:decode_breakdown} reports steady-state post-context throughput, excluding prefill to isolate recurring routing and index-preparation costs.

Decode-only gains increase from 7.67\% at 128 output tokens to 13.14\% at 8192 tokens. As the K-summary view grows, routing exposes more latency on the synchronous attention path; preparing indices early removes this wait from target attention.

\begin{figure}[t]
    \centering
    \includegraphics[width=0.9\linewidth]{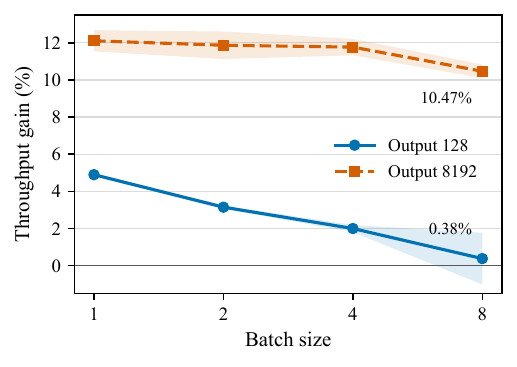}
    \caption{End-to-end throughput gain of ES-RocketKV over RocketKV on 8K-token inputs across batch sizes. Shaded regions show sample standard deviations over three runs.}
    \label{fig:runtime_scope}
\end{figure}

Figure~\ref{fig:runtime_scope} tests whether this advantage persists under batching. From batch 1 to 8, the gain remains 10.47--12.11\% for 8192-token outputs but falls from 4.90\% to 0.38\% for 128-token outputs, with the final mean below its run-to-run variation. Batching therefore preserves the benefit for long decodes, while short requests offer fewer steps over which to amortize fixed costs.

\begin{table}[t]
\centering
\small
\setlength{\tabcolsep}{4pt}
\renewcommand{\arraystretch}{1.08}
\begin{tabular}{@{}p{0.29\columnwidth}p{0.65\columnwidth}@{}}
\toprule
Item & Setting \\
\midrule
Method defaults & ES-Mid; layer 0 dense; local tail outside routing \\
Proxy projection & Offline SVD of \(W_q^{\ell+1}\); rank-128 BF16 for CPU routing \\
Runtime configuration & 40K main and 8K batch-scaling inputs; Stage-1: 4096; \(p=4\), \(k=32\), \(pk=128\); outputs 128--8192 \\
Backend / hardware & TensorRT-LLM 1.3.0rc15 with CUDA Graphs; RTX 6000D; Xeon Platinum 8481C \\
Routing offload & Routing and FP8 K-summary on CPU; KV cache on GPU; no prefetch \\
Measurement & 40K: one run; 8K: three runs, mean $\pm$ s.d. \\
Memory accounting & KV/K-summary allocation only; weights and workspaces excluded \\
\bottomrule
\end{tabular}
\caption{Key reproducibility settings for EarlySelectKV.}
\label{tab:reproducibility_settings}
\end{table}

\section{Reproducibility}
\label{app:reproducibility}
Within each quality comparison, the base and ES variants use identical prompts, decoding settings, and KV-access budgets. Profiling reuses the same prompt subsets across anchors, and runtime comparisons hold the sparse-attention configuration fixed. Table~\ref{tab:reproducibility_settings} lists the remaining shared settings.

\begin{table*}[t]
\centering
\scriptsize
\setlength{\tabcolsep}{2.0pt}
\renewcommand{\arraystretch}{1.05}
\resizebox{\textwidth}{!}{%
\begin{tabular}{c c cccccccccccccccc}
\toprule
Budget & Method & \rotatebox{45}{\strut NarrativeQA} & \rotatebox{45}{\strut QASPER} & \rotatebox{45}{\strut MultiFieldQA} & \rotatebox{45}{\strut HotpotQA} & \rotatebox{45}{\strut 2WikiMQA} & \rotatebox{45}{\strut MuSiQue} & \rotatebox{45}{\strut GovReport} & \rotatebox{45}{\strut QMSum} & \rotatebox{45}{\strut MultiNews} & \rotatebox{45}{\strut TREC} & \rotatebox{45}{\strut TriviaQA} & \rotatebox{45}{\strut SamSum} & \rotatebox{45}{\strut PassageRetrieval} & \rotatebox{45}{\strut PassageCount} & \rotatebox{45}{\strut LCC} & \rotatebox{45}{\strut RepoBench} \\
\midrule
\rowcolor{gray!15}\multicolumn{18}{c}{\textit{Llama-3-8B-Instruct}} \\
FULL & FullCache & 21.17 & 43.17 & 46.54 & 46.81 & 39.93 & 23.05 & 30.94 & 22.70 & 27.97 & 73.00 & 90.00 & 42.16 & 66.50 & 8.00 & 60.63 & 49.07 \\
\midrule
64 & TopK & 19.97 & 42.53 & 46.29 & 47.32 & 39.90 & 22.75 & 29.20 & 22.37 & 27.10 & 72.50 & 90.36 & 42.16 & 67.00 & 7.50 & 64.88 & 57.64 \\
 & RocketKV & 18.44 & \textbf{30.47} & \textbf{46.57} & \textbf{46.31} & \textbf{34.83} & 21.34 & \textbf{20.72} & \textbf{21.14} & \textbf{23.82} & 55.00 & \textbf{89.68} & 38.20 & \textbf{65.50} & \textbf{6.00} & 62.36 & 57.92 \\
 & ES-RocketKV & \textbf{19.19} & 28.77 & 46.42 & 44.88 & 34.78 & \textbf{21.47} & 20.62 & 21.09 & 23.72 & \textbf{57.00} & 89.49 & \textbf{38.66} & \textbf{65.50} & 5.50 & \textbf{64.22} & \textbf{58.16} \\
\midrule
128 & TopK & 20.55 & 43.15 & 45.16 & 46.81 & 41.52 & 22.12 & 29.19 & 22.42 & 27.45 & 72.50 & 90.86 & 42.17 & 66.50 & 8.50 & 64.76 & 55.91 \\
 & RocketKV & 18.15 & 40.40 & 45.92 & \textbf{46.28} & 36.89 & 21.92 & 21.75 & 21.13 & \textbf{24.19} & \textbf{66.50} & \textbf{90.05} & \textbf{40.86} & 66.50 & \textbf{5.00} & \textbf{63.18} & \textbf{58.11} \\
 & ES-RocketKV & \textbf{18.77} & \textbf{41.41} & \textbf{46.15} & 46.16 & \textbf{38.23} & \textbf{22.27} & \textbf{22.13} & \textbf{21.27} & 24.03 & \textbf{66.50} & 89.12 & 40.10 & \textbf{67.00} & \textbf{5.00} & 63.09 & 57.26 \\
\midrule
256 & TopK & 21.15 & 43.12 & 45.72 & 46.62 & 41.13 & 22.70 & 29.93 & 22.32 & 27.42 & 72.50 & 91.00 & 42.29 & 66.50 & 8.00 & 64.12 & 52.73 \\
 & RocketKV & 19.37 & \textbf{41.84} & 45.89 & 46.43 & 38.82 & 22.91 & 23.74 & \textbf{21.70} & 24.54 & \textbf{70.50} & 89.36 & 41.68 & \textbf{67.00} & 6.00 & 63.46 & 55.64 \\
 & ES-RocketKV & \textbf{19.75} & 41.57 & \textbf{46.68} & \textbf{46.79} & \textbf{39.19} & \textbf{23.14} & \textbf{23.82} & 21.49 & \textbf{24.58} & \textbf{70.50} & \textbf{89.87} & \textbf{42.34} & 66.50 & \textbf{7.00} & \textbf{64.21} & \textbf{56.12} \\
\midrule
512 & TopK & 21.49 & 43.51 & 45.29 & 46.91 & 41.38 & 22.72 & 30.37 & 22.67 & 27.59 & 72.50 & 91.00 & 42.36 & 66.50 & 8.00 & 62.43 & 51.34 \\
 & RocketKV & \textbf{19.97} & \textbf{43.00} & \textbf{46.36} & 46.93 & \textbf{40.21} & \textbf{23.08} & \textbf{26.75} & \textbf{22.31} & \textbf{26.05} & \textbf{71.50} & 89.95 & \textbf{42.84} & 66.50 & \textbf{7.00} & 63.35 & 53.38 \\
 & ES-RocketKV & 19.51 & 42.90 & 45.96 & \textbf{47.31} & 39.86 & 22.75 & 26.31 & 22.14 & 25.90 & \textbf{71.50} & \textbf{90.08} & 42.76 & \textbf{67.00} & 6.50 & \textbf{63.67} & \textbf{54.41} \\
\midrule
1024 & TopK & 21.00 & 43.37 & 45.97 & 47.20 & 41.15 & 22.79 & 30.63 & 22.90 & 27.70 & 72.50 & 91.00 & 41.84 & 66.50 & 7.50 & 61.23 & 49.67 \\
 & RocketKV & \textbf{20.92} & \textbf{43.45} & \textbf{45.98} & \textbf{46.63} & 40.21 & \textbf{23.09} & \textbf{28.67} & 22.10 & \textbf{26.90} & \textbf{73.00} & \textbf{90.00} & \textbf{41.97} & \textbf{66.50} & \textbf{8.00} & 62.12 & 51.35 \\
 & ES-RocketKV & 20.85 & 43.30 & 45.56 & \textbf{46.63} & \textbf{40.65} & \textbf{23.09} & 28.60 & \textbf{22.13} & 26.78 & \textbf{73.00} & \textbf{90.00} & 41.96 & \textbf{66.50} & 7.50 & \textbf{62.22} & \textbf{52.52} \\
\midrule
2048 & TopK & 21.23 & 43.09 & 46.60 & 46.97 & 40.83 & 22.77 & 30.77 & 22.91 & 27.80 & 72.50 & 91.00 & 41.91 & 66.50 & 8.00 & 60.86 & 49.17 \\
 & RocketKV & 21.24 & \textbf{43.67} & 46.33 & 46.18 & \textbf{40.41} & 23.01 & \textbf{30.05} & \textbf{22.57} & 27.61 & \textbf{73.00} & \textbf{90.00} & \textbf{42.41} & \textbf{66.50} & \textbf{8.50} & 61.16 & \textbf{50.45} \\
 & ES-RocketKV & \textbf{21.33} & 43.29 & \textbf{46.53} & \textbf{46.68} & \textbf{40.41} & \textbf{23.05} & 29.72 & 22.50 & \textbf{27.64} & \textbf{73.00} & \textbf{90.00} & 41.75 & \textbf{66.50} & \textbf{8.50} & \textbf{61.45} & 50.25 \\
\midrule
4096 & TopK & 21.18 & 43.32 & 46.59 & 47.05 & 41.04 & 22.79 & 31.18 & 22.74 & 27.90 & 72.50 & 90.83 & 42.33 & 66.50 & 8.00 & 60.92 & 48.96 \\
 & RocketKV & \textbf{21.29} & \textbf{43.74} & 46.32 & \textbf{46.80} & 39.93 & \textbf{22.98} & 30.50 & \textbf{22.91} & 27.80 & \textbf{73.00} & \textbf{90.00} & 41.70 & \textbf{66.50} & \textbf{8.00} & 60.68 & \textbf{49.61} \\
 & ES-RocketKV & 21.25 & 43.33 & \textbf{46.50} & 46.47 & \textbf{40.48} & \textbf{22.98} & \textbf{30.52} & 22.86 & \textbf{27.83} & \textbf{73.00} & \textbf{90.00} & \textbf{42.40} & \textbf{66.50} & \textbf{8.00} & \textbf{60.72} & 49.17 \\
\bottomrule
\end{tabular}
}%
\caption{Task-level LongBench scores for Llama-3-8B-Instruct.}
\label{tab:longbench_task_llama3_8b_instruct}
\end{table*}

\begin{table*}[t]
\centering
\scriptsize
\setlength{\tabcolsep}{2.0pt}
\renewcommand{\arraystretch}{1.05}
\resizebox{\textwidth}{!}{%
\begin{tabular}{c c cccccccccccccccc}
\toprule
Budget & Method & \rotatebox{45}{\strut NarrativeQA} & \rotatebox{45}{\strut QASPER} & \rotatebox{45}{\strut MultiFieldQA} & \rotatebox{45}{\strut HotpotQA} & \rotatebox{45}{\strut 2WikiMQA} & \rotatebox{45}{\strut MuSiQue} & \rotatebox{45}{\strut GovReport} & \rotatebox{45}{\strut QMSum} & \rotatebox{45}{\strut MultiNews} & \rotatebox{45}{\strut TREC} & \rotatebox{45}{\strut TriviaQA} & \rotatebox{45}{\strut SamSum} & \rotatebox{45}{\strut PassageRetrieval} & \rotatebox{45}{\strut PassageCount} & \rotatebox{45}{\strut LCC} & \rotatebox{45}{\strut RepoBench} \\
\midrule
\rowcolor{gray!15}\multicolumn{18}{c}{\textit{Llama-3.1-8B-Instruct}} \\
FULL & FullCache & 32.12 & 46.58 & 55.73 & 57.84 & 48.61 & 30.82 & 34.78 & 25.42 & 27.07 & 73.00 & 92.14 & 43.76 & 99.50 & 7.33 & 65.36 & 55.42 \\
\midrule
64 & TopK & 31.35 & 45.44 & 55.80 & 58.40 & 48.92 & 31.56 & 33.73 & 24.41 & 26.28 & 73.00 & 92.25 & 43.06 & 99.50 & 7.92 & 65.12 & 53.72 \\
 & RocketKV & 27.17 & 29.99 & \textbf{57.32} & 56.65 & \textbf{44.88} & \textbf{28.77} & 22.85 & \textbf{23.24} & 23.30 & 56.50 & 91.07 & \textbf{40.10} & \textbf{99.00} & \textbf{9.00} & \textbf{63.50} & \textbf{52.23} \\
 & ES-RocketKV & \textbf{27.35} & \textbf{31.46} & 56.42 & \textbf{56.87} & 44.19 & 27.29 & \textbf{23.17} & 23.15 & \textbf{23.37} & \textbf{58.00} & \textbf{91.70} & 39.98 & 98.50 & 7.83 & 63.20 & 51.39 \\
\midrule
128 & TopK & 30.99 & 45.27 & 55.84 & 58.07 & 49.03 & 31.84 & 34.25 & 25.57 & 26.56 & 73.00 & 92.25 & 44.23 & 99.50 & 7.58 & 65.83 & 54.78 \\
 & RocketKV & \textbf{30.26} & \textbf{41.12} & \textbf{57.26} & 57.15 & \textbf{47.93} & \textbf{29.85} & \textbf{24.10} & 23.70 & 23.84 & \textbf{67.00} & 91.73 & \textbf{41.55} & \textbf{99.50} & 6.83 & \textbf{65.16} & 54.12 \\
 & ES-RocketKV & 29.76 & 40.68 & 56.31 & \textbf{58.17} & 46.76 & 29.29 & 23.97 & \textbf{24.25} & \textbf{24.07} & \textbf{67.00} & \textbf{91.75} & 41.17 & \textbf{99.50} & \textbf{7.90} & 64.55 & \textbf{54.36} \\
\midrule
256 & TopK & 30.01 & 45.38 & 55.33 & 57.69 & 49.64 & 32.20 & 34.68 & 25.54 & 26.75 & 73.00 & 91.95 & 44.24 & 99.50 & 8.00 & 66.04 & 55.52 \\
 & RocketKV & 30.54 & \textbf{44.03} & \textbf{56.22} & 57.28 & \textbf{47.61} & 30.22 & 26.12 & 24.52 & \textbf{24.42} & \textbf{71.00} & 91.78 & 42.55 & \textbf{100.00} & 7.00 & 65.06 & \textbf{55.41} \\
 & ES-RocketKV & \textbf{30.69} & 42.35 & 55.45 & \textbf{57.99} & 47.28 & \textbf{31.15} & \textbf{26.59} & \textbf{24.65} & 24.25 & 70.50 & \textbf{91.81} & \textbf{42.58} & 99.50 & \textbf{8.00} & \textbf{65.63} & 55.21 \\
\midrule
512 & TopK & 30.29 & 46.12 & 55.40 & 57.58 & 49.22 & 32.20 & 34.50 & 25.37 & 26.97 & 73.00 & 91.97 & 43.63 & 99.50 & 7.92 & 65.47 & 55.42 \\
 & RocketKV & 31.60 & \textbf{45.99} & 55.73 & \textbf{58.40} & \textbf{48.28} & \textbf{30.18} & 29.85 & \textbf{25.03} & \textbf{25.51} & \textbf{72.50} & 91.64 & \textbf{43.46} & \textbf{99.50} & \textbf{7.00} & \textbf{66.06} & \textbf{55.96} \\
 & ES-RocketKV & \textbf{31.75} & 45.49 & \textbf{55.89} & 58.18 & 48.08 & 29.77 & \textbf{30.02} & 24.68 & 25.48 & \textbf{72.50} & \textbf{92.11} & 43.03 & \textbf{99.50} & \textbf{7.00} & 65.98 & 55.81 \\
\midrule
1024 & TopK & 30.67 & 46.26 & 55.72 & 57.73 & 49.22 & 32.04 & 34.75 & 25.68 & 27.01 & 73.00 & 91.97 & 43.13 & 99.50 & 7.50 & 65.21 & 55.75 \\
 & RocketKV & \textbf{31.25} & \textbf{45.76} & 55.81 & 57.68 & \textbf{48.71} & 31.00 & \textbf{32.24} & \textbf{25.60} & \textbf{26.91} & \textbf{73.00} & 91.64 & 43.72 & \textbf{99.50} & \textbf{6.75} & 65.58 & 55.49 \\
 & ES-RocketKV & 31.16 & 45.22 & \textbf{55.87} & \textbf{57.78} & 48.68 & \textbf{32.02} & 32.17 & 25.48 & 26.66 & \textbf{73.00} & \textbf{91.97} & \textbf{43.89} & \textbf{99.50} & 6.50 & \textbf{66.13} & \textbf{55.97} \\
\midrule
2048 & TopK & 31.28 & 46.05 & 55.58 & 57.33 & 48.29 & 31.61 & 34.59 & 25.48 & 27.27 & 73.00 & 91.97 & 43.51 & 99.50 & 7.25 & 65.06 & 55.22 \\
 & RocketKV & 31.39 & 46.41 & \textbf{56.08} & 57.41 & 48.71 & 30.66 & \textbf{33.43} & \textbf{25.58} & 26.97 & \textbf{73.00} & 91.64 & \textbf{43.65} & \textbf{99.50} & \textbf{8.42} & \textbf{65.31} & 55.29 \\
 & ES-RocketKV & \textbf{31.57} & \textbf{46.43} & 55.84 & \textbf{57.94} & \textbf{48.92} & \textbf{31.90} & 33.41 & 25.48 & \textbf{27.16} & \textbf{73.00} & \textbf{91.97} & 43.64 & \textbf{99.50} & 6.17 & 65.18 & \textbf{56.04} \\
\midrule
4096 & TopK & 31.61 & 46.33 & 55.82 & 57.58 & 48.68 & 32.05 & 34.49 & 25.39 & 27.34 & 73.00 & 91.97 & 43.50 & 99.50 & 8.00 & 64.97 & 55.42 \\
 & RocketKV & \textbf{31.67} & 46.53 & \textbf{56.05} & 57.54 & 48.52 & 31.55 & 34.02 & \textbf{25.20} & \textbf{27.27} & 72.50 & \textbf{91.97} & \textbf{43.89} & \textbf{99.50} & \textbf{8.52} & \textbf{65.17} & 55.46 \\
 & ES-RocketKV & 31.12 & \textbf{46.56} & 55.89 & \textbf{58.09} & \textbf{48.54} & \textbf{31.85} & \textbf{34.09} & 25.10 & 26.97 & \textbf{73.00} & \textbf{91.97} & 43.35 & \textbf{99.50} & 6.07 & 65.13 & \textbf{56.04} \\
\bottomrule
\end{tabular}
}%
\caption{Task-level LongBench scores for Llama-3.1-8B-Instruct.}
\label{tab:longbench_task_llama3_1_8b_instruct}
\end{table*}

\begin{table*}[t]
\centering
\scriptsize
\setlength{\tabcolsep}{2.0pt}
\renewcommand{\arraystretch}{1.05}
\resizebox{\textwidth}{!}{%
\begin{tabular}{c c cccccccccccccccc}
\toprule
Budget & Method & \rotatebox{45}{\strut NarrativeQA} & \rotatebox{45}{\strut QASPER} & \rotatebox{45}{\strut MultiFieldQA} & \rotatebox{45}{\strut HotpotQA} & \rotatebox{45}{\strut 2WikiMQA} & \rotatebox{45}{\strut MuSiQue} & \rotatebox{45}{\strut GovReport} & \rotatebox{45}{\strut QMSum} & \rotatebox{45}{\strut MultiNews} & \rotatebox{45}{\strut TREC} & \rotatebox{45}{\strut TriviaQA} & \rotatebox{45}{\strut SamSum} & \rotatebox{45}{\strut PassageRetrieval} & \rotatebox{45}{\strut PassageCount} & \rotatebox{45}{\strut LCC} & \rotatebox{45}{\strut RepoBench} \\
\midrule
\rowcolor{gray!15}\multicolumn{18}{c}{\textit{Mistral-7B-Instruct-v0.2}} \\
FULL & FullCache & 28.00 & 32.61 & 48.43 & 42.26 & 26.44 & 19.01 & 33.16 & 24.29 & 27.16 & 71.00 & 86.92 & 43.61 & 83.80 & 2.95 & 59.97 & 61.52 \\
\midrule
64 & TopK & 25.46 & 24.03 & 44.39 & 37.87 & 24.23 & 17.75 & 30.42 & 19.76 & 27.05 & 71.00 & 87.11 & 40.80 & 75.60 & 2.43 & 58.55 & 57.99 \\
 & RocketKV & 22.24 & 19.12 & \textbf{48.10} & \textbf{37.73} & \textbf{23.80} & 16.08 & 23.04 & 22.61 & 24.26 & \textbf{59.00} & \textbf{86.43} & 39.42 & \textbf{50.01} & \textbf{4.01} & \textbf{55.47} & \textbf{55.84} \\
 & ES-RocketKV & \textbf{22.58} & \textbf{19.71} & 47.56 & 36.95 & 22.90 & \textbf{16.29} & \textbf{23.41} & \textbf{22.70} & \textbf{24.72} & \textbf{59.00} & 86.41 & \textbf{39.94} & 43.52 & 3.99 & 55.03 & 55.30 \\
\midrule
128 & TopK & 25.75 & 30.06 & 48.21 & 39.28 & 25.44 & 17.88 & 31.68 & 21.86 & 27.13 & 71.00 & 87.27 & 42.95 & 82.37 & 2.53 & 59.21 & 59.62 \\
 & RocketKV & \textbf{25.17} & \textbf{26.43} & \textbf{48.86} & 38.10 & 23.90 & \textbf{18.47} & \textbf{23.22} & \textbf{23.17} & \textbf{24.41} & 68.50 & \textbf{87.61} & \textbf{42.00} & 58.62 & \textbf{3.61} & \textbf{57.78} & 57.92 \\
 & ES-RocketKV & 23.55 & 26.22 & 48.51 & \textbf{39.48} & \textbf{25.35} & 17.25 & 23.00 & 22.69 & 24.22 & \textbf{69.50} & 87.26 & 41.74 & \textbf{61.57} & 2.99 & 57.00 & \textbf{58.79} \\
\midrule
256 & TopK & 26.08 & 32.01 & 48.40 & 39.64 & 25.15 & 17.97 & 32.49 & 23.51 & 27.27 & 71.00 & 87.29 & 44.36 & 83.94 & 3.24 & 59.93 & 60.67 \\
 & RocketKV & 25.50 & 28.05 & \textbf{48.99} & 38.43 & 26.42 & \textbf{18.55} & \textbf{25.99} & \textbf{23.84} & 24.74 & \textbf{70.50} & 86.85 & \textbf{42.85} & 74.26 & \textbf{3.29} & \textbf{58.64} & \textbf{59.54} \\
 & ES-RocketKV & \textbf{25.79} & \textbf{28.81} & 48.78 & \textbf{38.51} & \textbf{26.74} & 18.44 & 25.59 & 23.83 & \textbf{24.79} & 70.00 & \textbf{86.93} & 42.24 & \textbf{75.31} & 2.75 & 57.72 & 59.44 \\
\midrule
512 & TopK & 26.99 & 31.54 & 48.89 & 41.22 & 25.32 & 18.84 & 33.12 & 23.91 & 27.53 & 71.00 & 87.08 & 44.40 & 84.51 & 3.59 & 59.49 & 60.65 \\
 & RocketKV & 26.46 & 30.97 & \textbf{48.77} & 40.34 & 25.84 & 18.44 & 28.95 & 23.87 & 25.96 & \textbf{70.50} & 87.08 & \textbf{43.25} & 77.93 & \textbf{3.21} & \textbf{59.74} & \textbf{60.12} \\
 & ES-RocketKV & \textbf{26.99} & \textbf{31.03} & 48.46 & \textbf{40.76} & \textbf{26.05} & \textbf{19.12} & \textbf{29.25} & \textbf{24.11} & \textbf{26.14} & \textbf{70.50} & \textbf{87.65} & 43.17 & \textbf{81.83} & 3.00 & 59.19 & 60.04 \\
\midrule
1024 & TopK & 27.38 & 31.39 & 49.18 & 41.49 & 25.93 & 19.08 & 33.44 & 23.92 & 27.29 & 71.00 & 87.09 & 44.48 & 83.76 & 3.32 & 59.11 & 60.74 \\
 & RocketKV & \textbf{27.45} & \textbf{31.67} & \textbf{48.03} & 41.90 & 25.41 & \textbf{18.96} & \textbf{31.47} & \textbf{24.37} & \textbf{26.98} & \textbf{71.00} & \textbf{87.08} & 43.38 & \textbf{82.18} & \textbf{3.41} & \textbf{59.31} & \textbf{60.72} \\
 & ES-RocketKV & 27.29 & 31.22 & 47.80 & \textbf{42.09} & \textbf{25.82} & \textbf{18.96} & 31.13 & 24.36 & 26.90 & \textbf{71.00} & \textbf{87.08} & \textbf{44.04} & 80.61 & 3.28 & 59.26 & 60.55 \\
\midrule
2048 & TopK & 27.57 & 32.09 & 48.49 & 41.90 & 25.49 & 19.05 & 33.24 & 24.07 & 27.21 & 71.00 & 87.04 & 43.98 & 84.01 & 3.02 & 59.16 & 60.95 \\
 & RocketKV & 27.23 & \textbf{31.74} & \textbf{48.70} & 41.66 & 25.32 & 18.91 & \textbf{32.63} & 24.49 & 27.12 & \textbf{71.00} & \textbf{87.13} & 43.49 & 83.29 & 2.97 & 59.51 & 60.91 \\
 & ES-RocketKV & \textbf{27.51} & 31.45 & 48.28 & \textbf{42.00} & \textbf{25.87} & \textbf{19.13} & 32.45 & \textbf{24.50} & \textbf{27.20} & \textbf{71.00} & 87.08 & \textbf{43.62} & \textbf{84.24} & \textbf{3.07} & \textbf{59.53} & \textbf{60.96} \\
\midrule
4096 & TopK & 27.28 & 32.68 & 48.57 & 41.95 & 25.60 & 19.16 & 33.45 & 24.18 & 27.15 & 71.00 & 86.86 & 43.96 & 83.51 & 2.81 & 59.21 & 61.47 \\
 & RocketKV & \textbf{27.32} & 32.06 & 48.31 & 42.10 & 26.06 & \textbf{19.06} & 33.25 & 24.56 & 27.38 & \textbf{71.00} & \textbf{87.15} & 43.51 & \textbf{83.83} & 2.94 & 59.29 & \textbf{61.00} \\
 & ES-RocketKV & 27.27 & \textbf{32.41} & \textbf{48.59} & \textbf{42.27} & \textbf{26.28} & 18.96 & \textbf{33.43} & \textbf{24.58} & \textbf{27.48} & \textbf{71.00} & \textbf{87.15} & \textbf{44.27} & 83.33 & \textbf{3.04} & \textbf{59.37} & 60.88 \\
\bottomrule
\end{tabular}
}%
\caption{Task-level LongBench scores for Mistral-7B-Instruct-v0.2.}
\label{tab:longbench_task_mistral_7b_instruct_v0_2}
\end{table*}

\begin{table*}[t]
\centering
\scriptsize
\setlength{\tabcolsep}{2.0pt}
\renewcommand{\arraystretch}{1.05}
\resizebox{\textwidth}{!}{%
\begin{tabular}{c c cccccccccccccccc}
\toprule
Budget & Method & \rotatebox{45}{\strut NarrativeQA} & \rotatebox{45}{\strut QASPER} & \rotatebox{45}{\strut MultiFieldQA} & \rotatebox{45}{\strut HotpotQA} & \rotatebox{45}{\strut 2WikiMQA} & \rotatebox{45}{\strut MuSiQue} & \rotatebox{45}{\strut GovReport} & \rotatebox{45}{\strut QMSum} & \rotatebox{45}{\strut MultiNews} & \rotatebox{45}{\strut TREC} & \rotatebox{45}{\strut TriviaQA} & \rotatebox{45}{\strut SamSum} & \rotatebox{45}{\strut PassageRetrieval} & \rotatebox{45}{\strut PassageCount} & \rotatebox{45}{\strut LCC} & \rotatebox{45}{\strut RepoBench} \\
\midrule
\rowcolor{gray!15}\multicolumn{18}{c}{\textit{Qwen2.5-7B-Instruct}} \\
FULL & FullCache & 27.77 & 45.33 & 51.63 & 58.00 & 46.13 & 29.88 & 33.51 & 24.28 & 25.81 & 73.00 & 88.41 & 47.02 & 100.00 & 9.50 & 59.86 & 64.92 \\
\midrule
64 & TopK & 28.90 & 43.05 & 48.84 & 54.16 & 43.16 & 27.74 & 29.42 & 23.25 & 24.73 & 70.50 & 87.62 & 46.04 & 96.09 & 9.00 & 59.26 & 61.96 \\
 & RocketKV & 20.88 & \textbf{29.82} & \textbf{50.17} & \textbf{50.90} & \textbf{39.19} & 24.96 & \textbf{21.37} & \textbf{21.41} & \textbf{20.77} & \textbf{50.50} & \textbf{83.07} & \textbf{40.87} & \textbf{86.58} & \textbf{8.00} & \textbf{50.80} & \textbf{50.79} \\
 & ES-RocketKV & \textbf{21.56} & 27.97 & 50.13 & 49.92 & 37.50 & \textbf{25.09} & 20.38 & 21.09 & 20.61 & 49.00 & 82.91 & 39.56 & 83.83 & 7.50 & 49.01 & 50.40 \\
\midrule
128 & TopK & 28.30 & 43.73 & 49.13 & 54.26 & 45.12 & 28.62 & 28.52 & 23.37 & 24.41 & 69.50 & 87.58 & 46.44 & 97.25 & 9.50 & 59.13 & 62.50 \\
 & RocketKV & 26.05 & \textbf{38.90} & \textbf{51.30} & 53.53 & \textbf{42.80} & \textbf{26.49} & \textbf{23.43} & \textbf{21.66} & \textbf{22.13} & \textbf{61.00} & 83.90 & 42.74 & \textbf{94.00} & 7.50 & \textbf{55.34} & \textbf{57.37} \\
 & ES-RocketKV & \textbf{26.16} & 38.76 & 50.98 & \textbf{53.62} & 42.25 & 26.06 & 22.92 & 21.42 & 21.70 & 60.25 & \textbf{84.81} & \textbf{43.14} & 93.38 & \textbf{9.00} & 54.01 & 56.57 \\
\midrule
256 & TopK & 28.37 & 42.76 & 48.64 & 54.91 & 45.45 & 28.30 & 29.30 & 23.61 & 24.03 & 70.50 & 86.14 & 46.01 & 98.50 & 9.50 & 58.76 & 63.04 \\
 & RocketKV & \textbf{28.65} & \textbf{41.72} & \textbf{51.71} & 54.43 & \textbf{43.76} & 27.43 & \textbf{25.37} & 22.29 & \textbf{22.20} & \textbf{64.50} & 85.31 & 44.47 & \textbf{95.42} & \textbf{7.50} & 58.32 & \textbf{60.37} \\
 & ES-RocketKV & 27.31 & 41.68 & 51.16 & \textbf{54.52} & 43.65 & \textbf{27.96} & 25.33 & \textbf{22.66} & 21.90 & \textbf{64.50} & \textbf{85.48} & \textbf{44.55} & 94.75 & \textbf{7.50} & \textbf{58.39} & 59.87 \\
\midrule
512 & TopK & 28.39 & 43.07 & 49.48 & 55.29 & 45.47 & 28.51 & 30.20 & 23.87 & 24.73 & 71.50 & 86.32 & 46.06 & 98.50 & 9.00 & 58.78 & 63.43 \\
 & RocketKV & \textbf{27.81} & \textbf{42.40} & \textbf{49.69} & 55.43 & 44.05 & 28.89 & 27.92 & \textbf{23.38} & 23.48 & \textbf{66.00} & \textbf{87.63} & 45.16 & \textbf{96.75} & \textbf{9.50} & \textbf{58.80} & 60.76 \\
 & ES-RocketKV & 27.60 & 42.27 & 49.64 & \textbf{55.83} & \textbf{44.24} & \textbf{29.23} & \textbf{28.16} & 23.17 & \textbf{23.65} & 65.50 & 87.38 & \textbf{45.78} & 96.25 & \textbf{9.50} & 58.50 & \textbf{60.78} \\
\midrule
1024 & TopK & 27.11 & 43.14 & 49.63 & 55.39 & 44.48 & 29.37 & 29.05 & 23.89 & 25.13 & 70.50 & 86.41 & 45.67 & 98.50 & 9.00 & 58.94 & 63.15 \\
 & RocketKV & 28.77 & 41.65 & 49.31 & 54.31 & 43.73 & \textbf{30.05} & 28.64 & 23.68 & \textbf{23.95} & \textbf{68.50} & \textbf{85.77} & \textbf{46.01} & \textbf{97.00} & \textbf{9.00} & \textbf{58.75} & \textbf{62.05} \\
 & ES-RocketKV & \textbf{29.18} & \textbf{41.70} & \textbf{49.85} & \textbf{55.06} & \textbf{43.77} & 30.00 & \textbf{28.75} & \textbf{23.91} & 23.90 & 67.50 & 85.74 & 45.55 & \textbf{97.00} & \textbf{9.00} & 58.63 & 61.74 \\
\midrule
2048 & TopK & 27.93 & 42.77 & 49.73 & 55.99 & 44.50 & 29.33 & 29.20 & 23.52 & 25.06 & 71.00 & 86.20 & 45.98 & 98.25 & 9.00 & 58.88 & 62.72 \\
 & RocketKV & \textbf{28.97} & \textbf{42.38} & 49.40 & 54.45 & \textbf{44.66} & 29.26 & \textbf{30.06} & 23.93 & \textbf{25.28} & \textbf{69.50} & 85.53 & \textbf{45.94} & \textbf{98.00} & \textbf{9.00} & \textbf{58.84} & 62.68 \\
 & ES-RocketKV & 28.69 & 42.31 & \textbf{49.92} & \textbf{55.44} & 44.65 & \textbf{29.64} & 29.82 & \textbf{24.10} & 24.91 & 69.00 & \textbf{85.57} & 45.44 & \textbf{98.00} & \textbf{9.00} & 58.46 & \textbf{62.92} \\
\midrule
4096 & TopK & 27.82 & 42.58 & 48.93 & 56.40 & 45.12 & 29.75 & 30.20 & 23.85 & 24.86 & 71.00 & 86.40 & 46.22 & 97.75 & 9.00 & 58.98 & 62.78 \\
 & RocketKV & \textbf{27.85} & \textbf{42.76} & 49.37 & \textbf{55.53} & 43.94 & 29.52 & 29.25 & \textbf{24.12} & \textbf{25.18} & \textbf{70.50} & 85.80 & \textbf{45.78} & \textbf{97.75} & \textbf{9.00} & 58.81 & 62.60 \\
 & ES-RocketKV & 27.63 & 42.63 & \textbf{49.89} & 55.27 & \textbf{44.57} & \textbf{29.66} & \textbf{30.33} & 24.07 & 25.13 & 70.00 & \textbf{86.39} & 45.74 & \textbf{97.75} & \textbf{9.00} & \textbf{58.96} & \textbf{62.81} \\
\bottomrule
\end{tabular}
}%
\caption{Task-level LongBench scores for Qwen2.5-7B-Instruct.}
\label{tab:longbench_task_qwen2_5_7b_instruct}
\end{table*}

\FloatBarrier

\begin{table*}[t]
\centering
\small
\setlength{\tabcolsep}{5pt}
\renewcommand{\arraystretch}{1.08}

\begin{minipage}[t]{0.55\textwidth}
\centering
\begin{tabular}{l l c c}
\toprule
Model & Method & Acc.@64 & First 100\% Budget \\
\midrule
Mistral-7B-v0.2 & TopK     & 100.0 & 64  \\
                 & RocketKV & 100.0 & 64  \\
                 & ES-In    & 100.0 & 64  \\
                 & ES-Mid   & 100.0 & 64  \\
                 & ES-Mix   & 100.0 & 64  \\
\midrule
Llama-3-8B       & TopK     & 100.0 & 64  \\
                 & RocketKV & 98.0  & 128 \\
                 & ES-In    & 100.0 & 64  \\
                 & ES-Mid   & 98.0  & 128 \\
                 & ES-Mix    & 100.0 & 64  \\
\midrule
Llama-3.1-8B     & TopK     & 100.0 & 64  \\
                 & RocketKV & 88.0  & 128 \\
                 & ES-In    & 85.0  & --  \\
                 & ES-Mid   & 88.0  & 128 \\
                 & ES-Mix   & 86.0 & 128  \\
\bottomrule
\end{tabular}
\end{minipage}
\hfill
\begin{minipage}[t]{0.40\textwidth}
\centering
\begin{tabular}{c c c c c}
\toprule
\multicolumn{5}{c}{\textit{Qwen2.5-7B-Instruct}} \\
\midrule
Budget & TopK & RocketKV & ES-In & ES-Mid \\
\midrule
64   & 76.0 & 48.0 & \textbf{56.0} & 48.0 \\
128  & 76.0 & 64.0 & \textbf{67.0} & 66.0 \\
256  & 74.0 & \textbf{81.0} & 79.0 & 80.0 \\
512  & 75.0 & 76.0 & 77.0 & \textbf{78.0} \\
1024 & 76.0 & \textbf{78.0} & 76.0 & 76.0 \\
2048 & 75.0 & 76.0 & 76.0 & \textbf{77.0} \\
4096 & 74.0 & \textbf{77.0} & 75.0 & 76.0 \\
\bottomrule
\end{tabular}
\end{minipage}

\caption{
NIAH/Passkey exact matching accuracy.
All settings use \texttt{10x10x3\_7digits}; context lengths are 4096 words for Llama-3-8B-Instruct, 20480 words for Mistral-7B-Instruct-v0.2, and 81920 words for Llama-3.1-8B-Instruct and Qwen2.5-7B-Instruct.
For compactness, models reaching perfect retrieval are summarized by the lowest budget achieving 100\% accuracy, while Qwen2.5-7B-Instruct is shown separately.
}
\label{tab:niah_full}
\end{table*}

\FloatBarrier

\begin{table*}[t]
\centering
\small
\setlength{\tabcolsep}{5pt}
\renewcommand{\arraystretch}{1.08}
\begin{tabular}{c c c c c c c}
\toprule
Context & Model & Budget & FullCache & TopK & RocketKV & ES-RocketKV \\
\midrule
RULER-8K & Mistral-7B-Instruct-v0.2 & 128 & 90.65 & 89.52 & \textbf{81.33} & 81.16 \\
 &  & 256 & 90.65 & 90.34 & \textbf{85.09} & 84.95 \\
 &  & 512 & 90.65 & 90.45 & \textbf{88.53} & 88.50 \\
 &  & 1024 & 90.65 & 90.80 & \textbf{89.65} & 89.63 \\
 &  & 2048 & 90.65 & 90.61 & 89.70 & \textbf{90.07} \\
 & Llama-3.1-8B-Instruct & 128 & 95.31 & 92.71 & \textbf{85.62} & 85.13 \\
 &  & 256 & 95.31 & 94.15 & \textbf{89.72} & 89.57 \\
 &  & 512 & 95.31 & 94.87 & \textbf{91.88} & 91.62 \\
 &  & 1024 & 95.31 & 95.17 & \textbf{94.02} & 93.64 \\
 &  & 2048 & 95.31 & 95.27 & \textbf{94.67} & 94.45 \\
 & Qwen2.5-7B-Instruct & 128 & 93.36 & 73.71 & \textbf{70.45} & 69.17 \\
 &  & 256 & 93.36 & 75.19 & \textbf{74.41} & 73.09 \\
 &  & 512 & 93.36 & 75.89 & \textbf{74.06} & 73.70 \\
 &  & 1024 & 93.36 & 75.91 & 74.75 & \textbf{75.31} \\
 &  & 2048 & 93.36 & 75.80 & 74.17 & \textbf{75.21} \\
\midrule
RULER-16K & Mistral-7B-Instruct-v0.2 & 128 & 86.23 & 83.48 & 75.18 & \textbf{75.89} \\
 &  & 256 & 86.23 & 85.23 & 78.27 & \textbf{80.38} \\
 &  & 512 & 86.23 & 85.65 & 82.72 & \textbf{83.05} \\
 &  & 1024 & 86.23 & 86.26 & \textbf{84.44} & 84.34 \\
 &  & 2048 & 86.23 & 86.49 & 85.33 & \textbf{85.49} \\
 & Llama-3.1-8B-Instruct & 128 & 93.22 & 90.31 & \textbf{85.09} & 84.70 \\
 &  & 256 & 93.22 & 91.56 & \textbf{87.75} & 87.29 \\
 &  & 512 & 93.22 & 92.75 & 89.52 & \textbf{89.61} \\
 &  & 1024 & 93.22 & 93.02 & \textbf{91.44} & 91.38 \\
 &  & 2048 & 93.22 & 93.16 & 92.11 & \textbf{92.22} \\
 & Qwen2.5-7B-Instruct & 128 & 92.03 & 68.16 & \textbf{63.75} & 63.57 \\
 &  & 256 & 92.03 & 69.91 & \textbf{67.27} & 66.78 \\
 &  & 512 & 92.03 & 70.16 & 68.73 & \textbf{69.67} \\
 &  & 1024 & 92.03 & 70.74 & \textbf{70.54} & 70.05 \\
 &  & 2048 & 92.03 & 71.68 & 70.66 & \textbf{71.14} \\
\midrule
RULER-32K & Mistral-7B-Instruct-v0.2 & 128 & 79.74 & 74.26 & \textbf{65.49} & 65.33 \\
 &  & 256 & 79.74 & 76.67 & 72.27 & \textbf{72.35} \\
 &  & 512 & 79.74 & 79.26 & 74.47 & \textbf{74.56} \\
 &  & 1024 & 79.74 & 79.87 & 75.88 & \textbf{77.08} \\
 &  & 2048 & 79.74 & 80.61 & \textbf{77.99} & 77.85 \\
 & Llama-3.1-8B-Instruct & 128 & 88.43 & 86.17 & 82.73 & \textbf{82.98} \\
 &  & 256 & 88.43 & 87.23 & 84.59 & \textbf{84.95} \\
 &  & 512 & 88.43 & 88.27 & 85.61 & \textbf{85.77} \\
 &  & 1024 & 88.43 & 88.83 & \textbf{86.32} & 86.09 \\
 &  & 2048 & 88.43 & 88.36 & \textbf{87.56} & 87.34 \\
 & Qwen2.5-7B-Instruct & 128 & 89.37 & 64.01 & \textbf{55.71} & 55.18 \\
 &  & 256 & 89.37 & 64.96 & \textbf{60.91} & 60.80 \\
 &  & 512 & 89.37 & 64.43 & 62.22 & \textbf{62.82} \\
 &  & 1024 & 89.37 & 64.35 & \textbf{64.89} & 64.05 \\
 &  & 2048 & 89.37 & 65.54 & \textbf{64.02} & 62.97 \\
\bottomrule
\end{tabular}
\caption{Average matching accuracy (\%) on RULER across context lengths.}
\label{tab:ruler_average}
\end{table*}

\FloatBarrier

\begin{table*}[t]
\centering
\small
\setlength{\tabcolsep}{5pt}
\renewcommand{\arraystretch}{1.08}
\begin{tabular}{c c c c c c}
\toprule
Model & Budget & RocketKV & ES-In & ES-Mid & ES-Mix \\
\midrule
Llama-3-8B-Instruct & 64 & 39.89 & 39.54 & 39.97 & \textbf{40.02} \\
 & 128 & 41.68 & 41.62 & 41.78 & \textbf{41.92} \\
 & 256 & 42.43 & 42.45 & \textbf{42.72} & 42.65 \\
 & 512 & 43.07 & 43.02 & 43.03 & \textbf{43.15} \\
 & 1024 & 43.18 & 43.16 & \textbf{43.21} & 43.14 \\
 & 2048 & 43.32 & 43.36 & 43.29 & \textbf{43.38} \\
 & 4096 & 43.23 & 43.27 & 43.25 & \textbf{43.29} \\
\midrule
Llama-3.1-8B-Instruct & 64 & 45.35 & 44.92 & 45.24 & \textbf{45.36} \\
 & 128 & \textbf{47.57} & 47.48 & 47.47 & 47.46 \\
 & 256 & 48.36 & 48.28 & 48.35 & \textbf{48.41} \\
 & 512 & 49.17 & 49.21 & 49.08 & \textbf{49.26} \\
 & 1024 & 49.41 & \textbf{49.62} & 49.50 & 49.47 \\
 & 2048 & 49.59 & \textbf{49.78} & 49.63 & 49.55 \\
 & 4096 & 49.71 & \textbf{49.82} & 49.58 & 49.58 \\
\midrule
Mistral-7B-Instruct-v0.2 & 64 & \textbf{36.70} & 35.62 & 36.25 & 36.25 \\
 & 128 & 39.24 & 39.11 & \textbf{39.32} & 39.24 \\
 & 256 & \textbf{41.03} & 40.73 & 40.98 & 40.94 \\
 & 512 & 41.96 & \textbf{42.39} & 42.33 & 42.18 \\
 & 1024 & 42.71 & 42.57 & 42.59 & \textbf{42.75} \\
 & 2048 & 42.88 & 42.82 & \textbf{42.99} & 42.86 \\
 & 4096 & 43.05 & \textbf{43.20} & 43.14 & 43.13 \\
\bottomrule
\end{tabular}
\caption{Average LongBench scores of RocketKV and EarlySelectKV routing variants.}
\label{tab:longbench_lookahead_variants}
\end{table*}

\begin{table*}[t]
\centering
\small
\setlength{\tabcolsep}{5pt}
\renewcommand{\arraystretch}{1.08}
\begin{tabular}{c c c c c c c}
\toprule
Context & Model & Budget & RocketKV & ES-In & ES-Mid & ES-Mix \\
\midrule
RULER-8K & Mistral-7B-Instruct-v0.2 & 128 & \textbf{81.33} & 80.72 & 81.16 & 80.73 \\
 &  & 256 & 85.09 & 84.76 & 84.95 & \textbf{85.18} \\
 &  & 512 & \textbf{88.53} & 87.88 & 88.50 & 88.31 \\
 &  & 1024 & \textbf{89.65} & 89.49 & 89.63 & 89.58 \\
 &  & 2048 & 89.70 & 89.96 & \textbf{90.07} & 89.77 \\
 & Llama-3.1-8B-Instruct & 128 & \textbf{85.62} & 84.54 & 85.13 & 85.38 \\
 &  & 256 & \textbf{89.72} & 88.67 & 89.57 & 89.71 \\
 &  & 512 & \textbf{91.88} & 91.06 & 91.62 & 91.59 \\
 &  & 1024 & \textbf{94.02} & 93.23 & 93.64 & 93.79 \\
 &  & 2048 & \textbf{94.67} & 94.39 & 94.45 & 94.49 \\
\midrule
RULER-16K & Mistral-7B-Instruct-v0.2 & 128 & 75.18 & 75.22 & \textbf{75.89} & 75.87 \\
 &  & 256 & 78.27 & 77.87 & \textbf{80.38} & 80.26 \\
 &  & 512 & 82.72 & 81.36 & \textbf{83.05} & 82.92 \\
 &  & 1024 & \textbf{84.44} & 83.66 & 84.34 & 84.05 \\
 &  & 2048 & 85.33 & 85.27 & \textbf{85.49} & 85.40 \\
 & Llama-3.1-8B-Instruct & 128 & \textbf{85.09} & 84.42 & 84.70 & 84.94 \\
 &  & 256 & \textbf{87.75} & 87.14 & 87.29 & 87.57 \\
 &  & 512 & 89.52 & 89.34 & \textbf{89.61} & 89.57 \\
 &  & 1024 & \textbf{91.44} & 91.28 & 91.38 & 91.28 \\
 &  & 2048 & 92.11 & 91.96 & 92.22 & \textbf{92.32} \\
\midrule
RULER-32K & Mistral-7B-Instruct-v0.2 & 128 & \textbf{65.49} & 64.19 & 65.33 & 65.04 \\
 &  & 256 & 72.27 & 71.85 & 72.35 & \textbf{72.57} \\
 &  & 512 & 74.47 & 74.49 & \textbf{74.56} & 74.54 \\
 &  & 1024 & 75.88 & 76.22 & \textbf{77.08} & 76.76 \\
 &  & 2048 & \textbf{77.99} & 77.29 & 77.85 & 77.50 \\
 & Llama-3.1-8B-Instruct & 128 & 82.73 & 81.71 & \textbf{82.98} & 82.87 \\
 &  & 256 & 84.59 & 84.58 & \textbf{84.95} & 84.90 \\
 &  & 512 & 85.61 & 85.56 & 85.77 & \textbf{85.96} \\
 &  & 1024 & 86.32 & 86.39 & 86.09 & \textbf{86.54} \\
 &  & 2048 & \textbf{87.56} & 87.29 & 87.34 & 87.36 \\
\bottomrule
\end{tabular}
\caption{Average matching accuracy (\%) of RocketKV and EarlySelectKV routing variants on RULER.}
\label{tab:ruler_lookahead_variants}
\end{table*}

\end{document}